\documentclass[11pt]{article}

\usepackage[preprint]{acl}

\usepackage{times}
\usepackage{latexsym}

\usepackage[T1]{fontenc}

\usepackage[utf8]{inputenc}

\usepackage{microtype}

\usepackage{inconsolata}

\usepackage{graphicx}

\usepackage{multirow, multicol}
\usepackage{array}      
\usepackage{booktabs}   
\usepackage{multirow}  
\usepackage{colortbl}   
\usepackage{xcolor}     
\usepackage{fontawesome5}
\usepackage{textgreek}
\usepackage{amsmath,amssymb,amsfonts}
\usepackage{graphicx}
\usepackage{array} 
\usepackage{float}    
\usepackage{enumitem}
\usepackage{lipsum}
\usepackage{multicol}
\usepackage{longtable}
\usepackage{multirow}
\usepackage{geometry}
\usepackage{tabularx}
\usepackage{makecell}
\usepackage{amsthm}
\usepackage{mathrsfs}
\usepackage[title]{appendix}
\usepackage{textcomp}
\usepackage{manyfoot}
\usepackage{hyperref}
\usepackage[most]{tcolorbox}

\usepackage{algorithm}
\usepackage[noend]{algpseudocode}

\algrenewcommand\algorithmicrequire{\textbf{Input:}}
\algrenewcommand\algorithmicensure{\textbf{Output:}}

\newtcolorbox{takeawaybox}{
  colback=blue!5,   
  colframe=blue!20,   
  boxrule=0.5pt,       
  arc=1mm,            
  left=2mm, right=2mm, 
  top=1mm, bottom=1mm,
  enhanced,
  breakable,             
}

\usepackage{listings}
\lstdefinestyle{promptstyle}{
    basicstyle=\ttfamily\scriptsize,
    breaklines=true,
    frame=shadowbox,
    numbers=none,
    xleftmargin=1em,
    framexleftmargin=1em,
    backgroundcolor=\color{gray!5}, 
}

\title{Understanding Chain-of-Thought in Large Language Models via Topological Data Analysis}

\author{
  \textbf{Chenghao Li\textsuperscript{1}},
  \textbf{Chaoning Zhang\textsuperscript{1}\thanks{Corresponding author}},
  \textbf{Yi Lu\textsuperscript{2}},
  \textbf{Shuxu Chen\textsuperscript{3}},
\\
  \textbf{Xudong Wang\textsuperscript{3}},
  \textbf{Jiaquan Zhang\textsuperscript{1}},
  \textbf{Zhicheng Wang\textsuperscript{1}},
  \textbf{Zhengxun Jin\textsuperscript{3}},
\\
  \textbf{Kuien Liu\textsuperscript{4}},
  \textbf{Sung-Ho Bae\textsuperscript{3}},
  \textbf{Guoqing Wang\textsuperscript{1}},
  \textbf{Yang Yang\textsuperscript{1}},
  \textbf{Heng Tao Shen\textsuperscript{5,1}}
\\
\\
  \textsuperscript{1}UESTC;
  \textsuperscript{2}CNU;
  \textsuperscript{3}KHU;
  \textsuperscript{4}CAS;
  \textsuperscript{5}Tongji Univ.
}

\begin{document}
\maketitle

\begin{abstract}
With the development of large language models (LLMs), particularly with the introduction of the long reasoning chain technique, the reasoning ability of LLMs in complex problem-solving has been significantly enhanced. While acknowledging the power of long reasoning chains, we cannot help but wonder: \textit{Why do different reasoning chains perform differently in reasoning? What components of the reasoning chains play a key role?} Existing studies mainly focus on evaluating reasoning chains from a functional perspective, with little attention paid to their structural mechanisms. To address this gap, this work is the first to analyze and evaluate the quality of the reasoning chain from a structural perspective.
We apply persistent homology from Topological Data Analysis (TDA) to map reasoning steps into semantic space, extract topological features, and analyze structural changes. These changes reveal semantic coherence, logical redundancy, and identify logical breaks and gaps. By calculating homology groups, we assess connectivity and redundancy at various scales, using barcode and persistence diagrams to quantify stability and consistency. Our results show that, 
under the full-graph (exploration) view, greater topological complexity correlates with higher accuracy. Conditioned on success, the final answer path tends to be topologically simpler, with fewer redundant cycles.
This work provides a perspective on reasoning chain quality assessment and offers guidance for future optimization.
\end{abstract}

\section{Introduction}

Existing research indicates that introducing long reasoning chains into LLMs can significantly enhance reasoning capabilities~\cite{wei2022chain,yao2023tree,besta2024graph, ning2023skeleton,zheng2023take,madaan2023self,zheng2023take}. Long reasoning chains enable the model to decompose complex problems into sequential subproblems, enhancing both accuracy and interpretability of the reasoning process~\citep{sun2025survey,he2025can,yeo2025demystifying}. Current research primarily focuses on enhancing reasoning chain generation through algorithm optimization, architectural improvements, and adjustments to training strategies~\citep{aggarwal2025l1,luo2025deconstructing,feng2025efficient,jin2025zero,zhang2025improve}. However, there is less emphasis on the systematic evaluation of chaining quality. This lack of systematic evaluation makes it difficult to objectively compare different chaining methods and limits the optimization direction of chaining. Among a few related works, \cite{jiang2025mme} evaluates the quality, efficiency, and robustness of reasoning chains using indicators such as accuracy and recall. \cite{gao2025benchmarking} proposes the 'reasoning granularity combination rule' to quantify the model's ability to handle complex reasoning in specific tasks. \cite{wang2025beyond} analyzes the relationship between the granularity of chaining and the generalization performance of the model and explores the performance of chaining training methods in terms of data sample efficiency. \cite{zhou2025landscape} develops a visualization tool to analyze the ability of LLMs' reasoning paths. 

Existing evaluation methods primarily emphasize quantitative indicators of reasoning chains in terms of robustness and efficiency, while systematic research on the structural characteristics of the reasoning chain itself remains lacking~\citep{he2025can,qu2025survey}. As an explicit expression of the model's reasoning process, the structure of the reasoning chain contains the model's organization of semantic and logical relationships~\citep{liu2025logic}. A deep understanding of these structures helps to reveal the model's reasoning mechanism and provide an analytical perspective and optimization path for improving reasoning ability.

To fill the gap above, this work proposes a systematic evaluation method for reasoning chains based on TDA \cite{munch2017user,chazal2021introduction} from the perspective of structural quality. Using a pre-trained language model, the textual reasoning process of a reasoning chain can be embedded into a high-dimensional semantic vector space, forming a point cloud with spatial structure. This point cloud structure makes TDA an ideal tool for analyzing the reasoning steps of reasoning chains \cite{zhou2022learning}, as it captures structural patterns in the semantic space without relying on dimensionality reduction, extracting topological features with structural significance from complex nonlinear data. This work extracts and quantifies topological features, such as connectivity, closed-loop structures, and high-dimensional voids, from the point cloud of reasoning steps across multiple scales, using persistent homology to analyze the quality of reasoning chains. In the semantic space, semantically similar sentences are typically mapped to relatively close positions. Thus, persistent homology can reveal the underlying patterns of semantic relationships by analyzing the evolution of the geometric and topological structures in the embedding space across scales. Persistent homology focuses on the structural relationships of data at different scales and is scale-invariant and shape-independent. In other words, it does not rely on specific distances or geometric forms. This enables effective analysis of semantic coherence, logical redundancy, and the effectiveness of information transmission in reasoning paths of various scales and structural forms, while also identifying potential issues such as logical leaps and semantic breaks during the reasoning process. Additionally, this work proposes quantification indicators based on the lifespan, complexity, and other aspects of the thinking chain through persistent diagrams and barcode diagrams. This work reveals the reasons behind the effectiveness of reasoning chains from a structural perspective and provides constructive directions for future development. 

Overall, our contributions are summarized as follows:

\begin{itemize}
    \item To the best of our knowledge, We present the first analysis of the structural characteristics of reasoning chains and provide a new enlightenment regarding the correlation between quality and topological structure.
    \item The study introduces persistent homology from TDA to analyze the reasoning chain structure. It extracts topological features and analyzes structural changes across multiple scales. This approach enables a detailed examination of the reasoning chain's semantic coherence and logical redundancy, offering a novel and comprehensive method for evaluating reasoning chain quality.
    \item Our findings reveal that a more intricate reasoning chain enhances the likelihood of identifying the correct solution early in the process. Successful reasoning outcomes typically converge on simpler, more logical paths, characterized by reduced redundancy and minimal loops.
\end{itemize}
\section{Related Work}
\label{sec:relat}

\subsection{Reasoning Chain Modeling in LLMs}
CoT prompting has become a key technique for improving LLMs' multi-step reasoning in tasks such as arithmetic and commonsense inference \cite{wei2022chain, yu2023towards, chu2023navigate}. Extensions like Self-Consistency CoT \cite{wang2022self}, KG-CoT \cite{zhao2024kg}, and Dynamic Prompt CoT \cite{lu2022dynamic} enhance reasoning robustness, knowledge integration, and adaptability. More recent frameworks introduce structured planning and search mechanisms, such as Plan-and-Solve \cite{wang2023plan}, Tree-of-Thoughts (ToT) \cite{yao2023tree}, and Algorithm-of-Thoughts (AoT) \cite{sel2023algorithm}. Graph-of-Thoughts (GoT) \cite{besta2024graph} further generalizes reasoning into a graph structure, enabling parallel and dynamic multi-path reasoning.

These developments reflect a shift toward more interpretable and structured reasoning in LLMs. In this work, we take a new step by introducing TDA to analyze the geometry of reasoning chains.

\subsection{Topological Data Analysis}
TDA \cite{wasserman2018topological} offers a powerful framework for capturing multi-scale structural patterns in high-dimensional data. Central to TDA is persistent homology \cite{carlsson2009topology, zomorodian2004computing}, which identifies stable topological features such as loops and holes, even under noise \cite{turkes2022effectiveness}. In NLP, TDA has been used to model semantic clusters and discourse structures \cite{chiang2007discover, wagner2012computational}, abstract high-dimensional text representations \cite{singh2007topological}, and support applications in visualization \cite{karlgren2014semantic}, legal reasoning \cite{savle2019topological}, and syntactic analysis \cite{port2018persistent}. Recent work further integrates TDA with deep models, such as TopoBERT \cite{rathore2023topobert}, to analyze latent structures in Transformer embeddings \cite{fitz2022shape}.

These studies show that TDA enables a structured understanding of language data, revealing hidden relationships in semantic spaces. A detailed review is provided in Appendix Related Work.
\section{Method}

\begin{figure*}
    \centering
    \includegraphics[width=\linewidth]{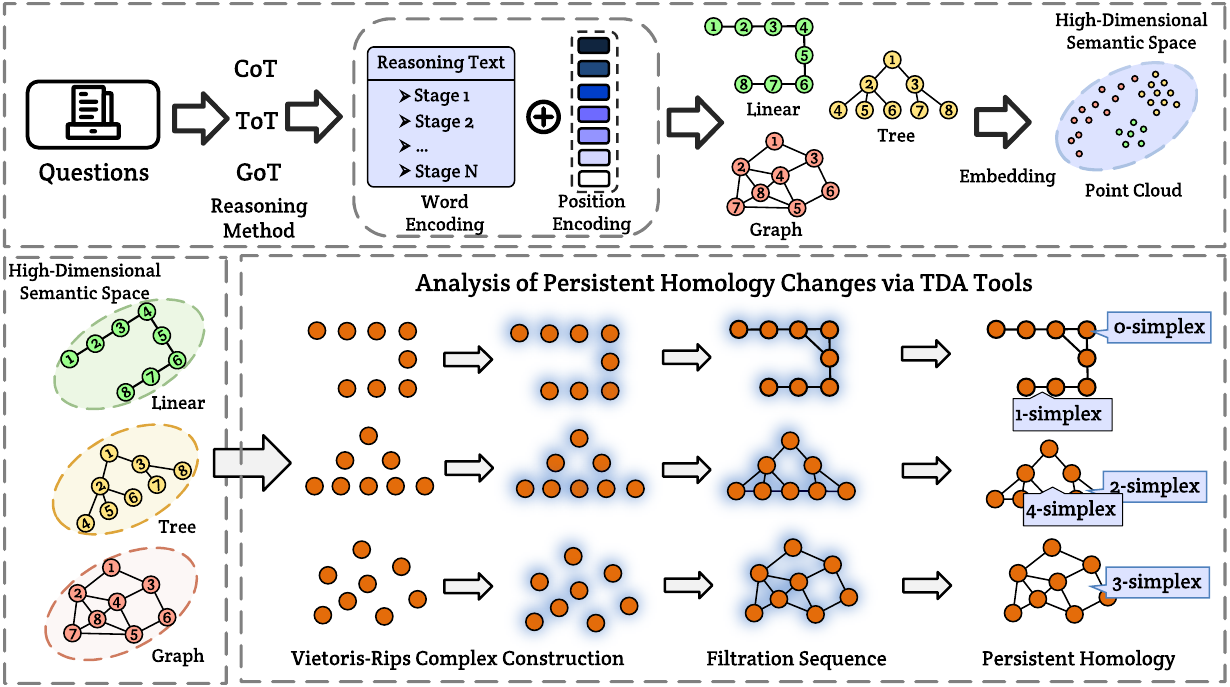}
    \caption{\textbf{Overall framework.} It presents the complete topological analysis pipeline from reasoning text to a semantic point cloud, then to a Vietoris--Rips complex, and finally to persistent homology (barcode/diagram).
}
    \label{fig:TDA}
\end{figure*}

Existing research on evaluating reasoning chains focuses on their performance. In contrast, this work analyzes the quality of reasoning chains from a structural perspective. In this section, we analyze reasoning chains with different structures. First, we map reasoning steps into a high-dimensional space using semantic embeddings and apply different positional encodings to capture the topological features of the reasoning structure (\S\ref{sec:cot_encoding}). Next, we construct a Vietoris-Rips complex (\S\ref{sec:vietoris_rips_complex}) and analyze the connectivity and redundancy of reasoning steps through homology group computation (\S\ref{sec:homology_group}), using persistent homology to extract topological features. Finally, we model barcode and persistence diagrams to quantify the lifecycle and distribution of topological features, revealing the stability and logical coherence of reasoning chains, providing quantitative metrics for reasoning quality assessment (\S\ref{sec:metrics&quantification}). Our method architecture diagram is shown in Fig.\ref{fig:TDA}.

\subsection{Geometric and Topological Representation of Reasoning Chains.}
\label{sec:cot_encoding}
In studying the reasoning mechanisms of large language models, a reasoning chain can be viewed as an explicit reasoning path generated by the model when solving complex tasks, consisting of a sequence of logically related linguistic steps. Formally, we denote a reasoning chain by
\begin{align}
S = (s_1, s_2, \dots, s_n),
\end{align}
where each reasoning step $s_i$ is a natural language sentence that may contain semantic content such as logical transitions, mathematical calculations, intermediate inferences, and hypothesis switching.
To investigate the structural properties of reasoning chains under the framework of TDA, we first need to embed these discrete linguistic steps into a high-dimensional semantic space that can capture geometric and topological relationships.

\noindent\textbf{Semantic Encoding: From Natural Language to High-Dimensional Vectors.}
Let $L$ be the set of natural language sentences. We introduce a semantic encoding function
\begin{align}
\Phi : L \to \mathbb{R}^d
\end{align}
which maps each reasoning step $s_i$ to a high-dimensional semantic vector
\begin{align}
\vec{x}_i = \Phi(s_i) \in \mathbb{R}^d,\quad i = 1,2,\dots,n.
\end{align}

This semantic embedding primarily reflects content similarity, i.e., semantic proximity. However, by itself it cannot directly encode the structural information of the reasoning chain, such as ordering, branching relations, and overall connectivity. Therefore, we additionally introduce structural positional encodings and combine them with the semantic vectors to explicitly capture the structure of reasoning.

Under different reasoning paradigms (CoT, ToT, GoT), reasoning chains exhibit linear, tree-like, or general graph-like structures, and the corresponding structural encoding schemes differ accordingly.

\noindent\textbf{CoT: Curvilinear Embedding of Linear Reasoning Chains.}
In the CoT paradigm, reasoning is viewed as a linearly progressing process:
\begin{align}
s_1 \rightarrow s_2 \rightarrow \dots \rightarrow s_n.
\end{align}

To encode this linear order structure, we adopt the classical sine--cosine positional encoding from Transformer architectures, denoted by $\mathrm{PE}(i)$, where $i$ is the position of the reasoning step in the sequence. The final embedding of each node is defined as
\begin{align}
\widetilde{\vec{x}}_i = \vec{x}_i + \mathrm{PE}(i),\quad i=1,2,\dots,n.
\end{align}

With this encoding, the point cloud of CoT embeddings in high-dimensional space typically appears as a “curve” lying on a low-dimensional manifold: it preserves semantic similarity while explicitly encoding the temporal order structure.

\noindent\textbf{ToT: Hierarchical Fan-Shaped Structures of Tree-Like Reasoning.}
ToT extends the reasoning chain into a tree structure, where each node has a well-defined hierarchical level and branching pattern. Let the depth of node $i$ be $d_i$, and let $b_i$ denote information about its branch or sibling nodes. To this end, we introduce two types of structural positional encodings:
\begin{align}
\mathrm{PE}_{\mathrm{depth}}(d_i), \qquad
\mathrm{PE}_{\mathrm{branch}}(b_i),
\end{align}
which encode the node's hierarchical depth and branch structure, respectively. The final embedding of node $i$ is defined as
\begin{align}
\widetilde{\vec{x}}_i
= \vec{x}_i
- \mathrm{PE}_{\mathrm{depth}}(d_i)
- \mathrm{PE}_{\mathrm{branch}}(b_i).
\end{align}

Under this encoding, the ToT embedding point cloud often exhibits a tree-like fan-shaped distribution: it radiates outward from the root node layer by layer, with different branches gradually separating in high-dimensional space, and the hierarchical structure manifests geometrically as a clear “radial” pattern.

\noindent\textbf{GoT: Graph Laplacian--Based Encoding of General Graph Structures.}
GoT has the strongest structural expressiveness, allowing the reasoning process to form a general graph structure. Unlike linear chains or trees, graph structures typically lack a fixed global order or a single hierarchy, so a more general graph-structure encoding is required.

To this end, we use graph Laplacian eigenvectors as structural positional encodings. Let $A$ be the adjacency matrix of the reasoning graph and $D$ the degree matrix. The (unnormalized) graph Laplacian of an undirected graph is
\begin{align}
L = D - A.
\end{align}

We perform eigen-decomposition on the (symmetric) Laplacian $L$:
\begin{align}
LU = U\Lambda,
\end{align}
where $U \in \mathbb{R}^{n \times d}$ stacks the first $d$ eigenvectors of $L$ (associated with the smallest eigenvalues), and
\begin{align}
\Lambda = \mathrm{diag}(\lambda_1,\lambda_2,\ldots,\lambda_d) \in \mathbb{R}^{d \times d}
\end{align}
is a diagonal matrix whose diagonal entries are the corresponding eigenvalues ($\lambda_1 \le \lambda_2 \le \cdots \le \lambda_d$).
The $i$-th row $U_i \in \mathbb{R}^d$ gives the Laplacian coordinates of node $i$. We define the graph-structural positional encoding as
\begin{align}
\mathrm{PE}_{\mathrm{graph}}(i) = U_i.
\end{align}

The final embedding of node $i$ in GoT is then given by
\begin{align}
\widetilde{\vec{x}}_i
= \vec{x}_i + \mathrm{PE}_{\mathrm{graph}}(i),\quad i=1,2,\dots,n.
\end{align}
With this representation, the GoT point cloud naturally reflects both the global connectivity and local neighborhood relations of the graph: nodes belonging to different subgraphs or communities form several relatively compact clusters in high-dimensional space, while key nodes (such as bridge nodes or nodes with high centrality) tend to lie in transitional regions between clusters.

\S\ref{subsec:semantic-embedding} provides a more detailed explanation of how reasoning processes are mapped into point-cloud representations.

\subsection{Topological Structural Analysis of Reasoning Chains}

In this section, we use TDA to analyze the changes in persistent homology in the structure of reasonig chains. 

\noindent\textbf{Reasoning Chains to Topology.}
\label{sec:vietoris_rips_complex}
Given the point-cloud representation of reasoning chains in the semantic space constructed in \S\ref{sec:cot_encoding}, we next build a Vietoris–Rips complex (\S\ref{sec:pre-vietoris_rips_complex}) to capture their underlying geometric and topological structure. Let \(X = \{\vec{x}_1,\ldots,\vec{x}_n\}\) denote the set of embedded reasoning steps (nodes) in a chain, tree, or graph of thought, and let \(d : X \times X \to \mathbb{R}_{\ge 0}\) be the Euclidean distance. For a scale parameter \(\epsilon > 0\), the Vietoris–Rips complex \(\mathrm{VR}_\epsilon(X)\) is defined as
\begin{equation}
\begin{split}
    \mathrm{VR}_\epsilon(X)
    = \bigl\{ \sigma \subseteq X \,\big|\, 
    \lVert \vec{x}_i - \vec{x}_j \rVert_2 \le \epsilon \\
    \text{for all } \vec{x}_i, \vec{x}_j \in \sigma \bigr\},
\end{split}
\end{equation}
where \(\sigma\) is a finite subset of points in \(X\). Thus, a collection of reasoning steps forms a simplex whenever all pairwise semantic distances between them are at most \(\epsilon\), and the resulting simplices encode local connectivity patterns within the reasoning process. By varying \(\epsilon\), we obtain a multi-scale family of complexes that reveals how local fragments of the reasoning chain gradually merge into global structures, providing the combinatorial backbone for the subsequent persistent homology analysis of reasoning quality.

\noindent\textbf{Multi-scale Reasoning Evolution Trajectory.}
We next track how the embedded step vectors in a reasoning chain gradually glue into larger semantic structures as the scale increases. Given the point cloud of reasoning steps, we consider a sequence of scale parameters
\(0 < \epsilon_0 < \epsilon_1 < \cdots < \epsilon_m\) and the corresponding Vietoris–Rips complexes:
\begin{equation}
    \mathrm{VR}_{\epsilon_0}(X)
    \subseteq \mathrm{VR}_{\epsilon_1}(X)
    \subseteq \cdots
    \subseteq \mathrm{VR}_{\epsilon_m}(X).
\end{equation}

As \(\epsilon\) grows, more edges and higher-order simplices are added whenever groups of reasoning steps become mutually close in the semantic space. The resulting nested complexes mirror how isolated steps, local subchains, and larger branches of the reasoning process gradually merge into a connected structure at coarser semantic resolutions. At each scale, both the vertex set and the simplices encode the connectivity pattern of the underlying reasoning chain.

For notational convenience, we write \(K_i := \mathrm{VR}_{\epsilon_i}(X)\), so the filtration can be expressed as
\begin{equation}
    K_0 \subset K_1 \subset \cdots \subset K_m,
\end{equation}
where each complex \(K_i\) corresponds to the reasoning geometry at scale \(\epsilon_i\).
This simplicial filtration serves as the input to our persistent homology pipeline, which summarizes how multi-scale semantic structures in the model’s reasoning (e.g., connected components and loops) are born and die across scales.

\noindent\textbf{Connected Components, Loops, and Cavities in the Reasoning Chains.} 
\label{sec:homology_group}
To analyze the topological features in reasoning chains, we compute the homology group for each scale complex. The homology group quantifies the topological properties of a space by measuring the structure of its components at different dimensions (\S\ref{sec:topological_features}).
Specifically, (1) the 0-dimensional homology group \( H_0 \) captures the connected components, reflecting how well the reasoning steps are semantically connected. (2) The 1-dimensional homology group \( H_1 \) identifies loops, representing logical redundancy or circular reasoning. (3) Higher-dimensional homology groups, such as the 2-dimensional group \( H_1 \), represent more complex structures like cavities, which suggest intricate reasoning patterns.

\noindent\textbf{Birth and Death of Topological Features.} In essence, this method extracts the topological features of the reasoning process by measuring the relationships between the elements at various levels of abstraction. These relationships help in identifying coherent, redundant, or complex patterns in the reasoning chain. The specific formula is shown in \S\ref{sec:homology_group_computation}.

To track the birth and death of topological features across scales, we use a series of nested complexes that create homology group mappings. These mappings are crucial for capturing the evolution of topological structures as $\epsilon$ changes.
\begin{equation}
    f_{i,j}\colon H_k\bigl(\mathrm{VR}_{\epsilon_i}(X)\bigr) \to H_k\bigl(\mathrm{VR}_{\epsilon_j}(X)\bigr),
    \text{for } i \le j
\end{equation}
where each mapping \(f_{i,j}\) is induced by the inclusion of complexes. As \(\epsilon\) increases, topological features like connected components and cycles emerge and then disappear. Specifically: (1) The birth scale \(b\) of a feature corresponds to the smallest \(\epsilon_i\) at which the feature appears in the homology group and did not exist in any preceding scale. (2) The death scale \(d\) is defined as the smallest \(\epsilon_j > b\) at which the feature either merges with another feature or becomes trivial, meaning it no longer exists as a distinct topological structure.
\begin{table}[t]
\centering
\small
\scalebox{0.9}{
\begin{tabular}{l l c c c}
\toprule
Dataset & Method & Acc. & $|H_0|$ & $|H_1|$ \\
\midrule
\multirow{3}{*}{GSM8K} 
  & CoT~\cite{wei2022chain} & 0.670 & 2.050 & 0.080 \\
  & ToT~\cite{yao2023tree} & 0.755 & 3.600 & 0.265 \\
  & GoT~\cite{besta2024graph} & 0.790 & 5.200 & 0.700 \\
\midrule
\multirow{3}{*}{MATH} 
  & CoT~\cite{wei2022chain} & 0.475 & 2.076 & 0.116 \\
  & ToT~\cite{yao2023tree} & 0.617 & 3.489 & 0.253 \\
  & GoT~\cite{besta2024graph} & 0.657 & 4.205 & 0.563 \\
\midrule
\multirow{3}{*}{MMLU} 
  & CoT~\cite{wei2022chain} & 0.529 & 2.070 & 0.043 \\
  & ToT~\cite{yao2023tree} & 0.541 & 3.825 & 0.280 \\
  & GoT~\cite{besta2024graph} & 0.579 & 6.450 & 0.928 \\
\bottomrule
\end{tabular}
}
\caption{\textbf{Accuracy and Betti numbers.} Comparing CoT/ToT/GoT across different datasets in terms of accuracy and $|H_0|$, $|H_1|$ (number of $H_0$ and $H_1$).}
\label{tab:tda_global}
\end{table}

\begin{table}[t]
\centering
\small
\scalebox{0.8}{
\begin{tabular}{l l c c c c}
\toprule
Dataset & Method 
& $H_0^{\max}$ & $H_0^{\mathrm{avg}}$
& $H_1^{\max}$ & $H_1^{\mathrm{avg}}$ \\
\midrule
\multirow{3}{*}{GSM8K} 
  & CoT~\cite{wei2022chain} & 0.030 & 0.019 & 0.003 & 0.002 \\
  & ToT~\cite{yao2023tree} & 0.046 & 0.030 & 0.040 & 0.024 \\
  & GoT~\cite{besta2024graph} & 0.060 & 0.040 & 0.058 & 0.026 \\
\midrule
\multirow{3}{*}{MATH} 
  & CoT~\cite{wei2022chain} & 0.031 & 0.020 & 0.003 & 0.002 \\
  & ToT~\cite{yao2023tree} & 0.045 & 0.029 & 0.043 & 0.026 \\
  & GoT~\cite{besta2024graph} & 0.052 & 0.034 & 0.063 & 0.027 \\
\midrule
\multirow{3}{*}{MMLU} 
  & CoT~\cite{wei2022chain} & 0.021 & 0.010 & 0.002 & 0.002 \\
  & ToT~\cite{yao2023tree} & 0.048 & 0.031 & 0.035 & 0.021 \\
  & GoT~\cite{besta2024graph} & 0.074 & 0.048 & 0.061 & 0.023 \\
\bottomrule
\end{tabular}
}
\caption{\textbf{Persistence statistics.} Max and average lifetimes) of $H_0$ and $H_1$ features.}
\label{tab:tda_persistence}
\end{table}

\section{Experiment}

\subsection{Metrics and Quantification}
\label{sec:metrics&quantification}

\noindent\textbf{Topological Structural Indicators.}
\label{sec:topological_structural_indicators}
In persistent homology analysis, the Barcode Diagram and Persistence Diagram are important tools used to visualize the birth and death of topological features~\cite{carlsson2009topology}. These diagrams not only provide an intuitive representation of the lifecycle of topological features in the data but also help quantitatively compare the topological structures across different datasets or reasoning processes. (The specific definitions of the Barcode Diagram and Persistence Diagram are given in the  \S\ref{sec:experimental_Metrics&quantification}.)

\noindent\textbf{Topological Feature Quantification} is essential for understanding the stability, complexity, and distribution of the underlying structure within the data~\cite{singh2007topological}. To achieve this, we utilize several key metrics that describe the persistence and distribution of topological features. Basic statistical measures, such as total lifetime length, average lifetime, and maximum lifetime, reflect the stability of the topological structure and the concentration of feature lifetimes. In addition, we quantify the complexity of the structure using topological feature count, and the persistence of these features is further captured through Persistent Entropy, which evaluates the distribution of topological structures across different scales. Together, these metrics provide a comprehensive assessment of the topological structure’s stability, complexity, and persistence. (\S\ref{sec:experimental_Metrics&quantification}).

\noindent\textbf{Setting.}
To analyze the performance of the reasoning chain, we employ the CoT~\cite{wei2022chain}, ToT~\cite{yao2023tree}, and GoT~\cite{besta2024graph} on the multi-step~\cite{cobbe2021gsm8k}, contest math~\cite{hendrycksmath2021},
and multi-subject reasoning~\cite{hendryckstest2021}. The problems cover various domains, such as algebra and geometry, and include detailed solutions. 

\subsection{Experimental Results and Analysis}

\noindent\textbf{Relationship between Structural Indicators and the Performance of Reasoning Chains.}
Tab.~\ref{tab:tda_global} and Tab.~\ref{tab:tda_persistence} report the reasoning performance of different chains across datasets and models, together with topological indicators such as $H_0$ and $H_1$. Overall, GoT consistently attains the highest accuracy, especially on more challenging tasks like those in the MATH dataset, where its richer topology—characterized by a larger number of $H_0$ components and more persistent $H_1$ loops—correlates with improved performance. In contrast, CoT follows an almost purely linear reasoning pattern, with the simplest topology and the lowest accuracy, particularly on tasks requiring deeper reasoning. ToT lies between the two: its branching structure is topologically more complex than that of CoT, but the absence of substantial merging and cyclic patterns still limits its effectiveness on complex problems compared with GoT. The performance gains of GoT are also more pronounced on MATH than on MMLU, suggesting that tasks demanding multi-step, rigorous reasoning benefit the most from GoT’s networked reasoning patterns. Moreover, different backbone models (3.5-turbo and 4o-mini; see Tab.~\ref{tab:tda_all_metrics} in \S\ref{sec:additional_experiments}) induce different levels of topological complexity for GoT, reflecting internal differences in their exploration strategies. 
Finally, these results demonstrate that TDA is a useful lens for evaluating reasoning structures: metrics such as $H_0$ and $H_1$ capture the transition from linear to branching to fully networked reasoning, in close alignment with empirical performance.

\begin{takeawaybox}
\textbf{Takeaway:} A more networked, topologically richer reasoning structure is strongly correlated with higher accuracy.
\end{takeawaybox}

\begin{figure*}
    \centering
    \includegraphics[width=\linewidth]{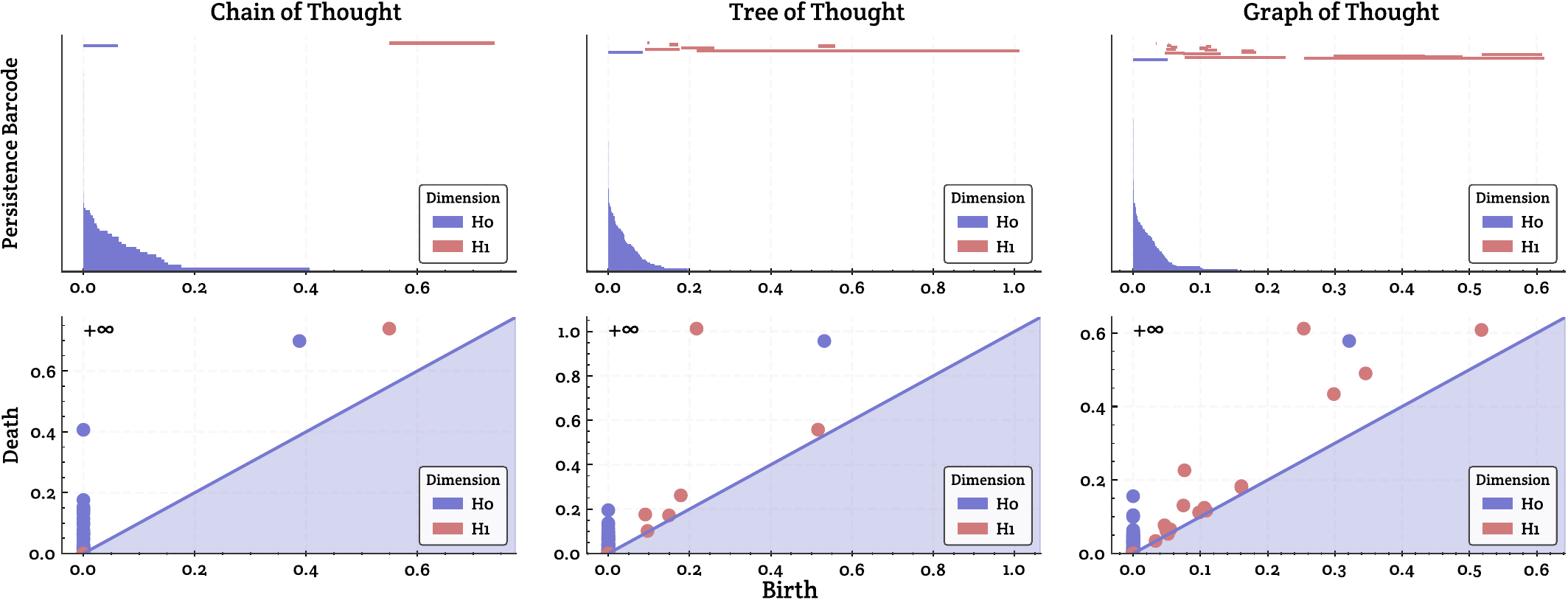}
    \caption{\textbf{Persistent barcode \& persistence diagram: CoT/ToT/GoT.} Birth–death patterns of connected components ($H_0$) and loops ($H_1$) across scales for CoT, ToT, and GoT. In each column, the top panel shows two-layer barcodes for $H_0$ and $H_1$ (horizontal bars), and the bottom panel shows a persistence diagram with a diagonal reference line, where points are colored by homology dimension and include $+\infty$ markers.
    }
    \label{persistence_barcode_diagram}
\end{figure*}

\noindent\textbf{Variation of Homological Features.} Based on the persistent homology visualizations in Fig.~\ref{persistence_barcode_diagram} (persistence barcodes and persistence diagrams), we can qualitatively compare the structural complexity and stability of different reasoning paradigms. Overall, CoT exhibits very few $H_1$ features, indicating that it rarely forms stable loop structures across scales and thus behaves more like a linear progression. Meanwhile, its $H_0$ evolution is relatively simple, suggesting that, as the filtration scale increases, the semantic representations merge into a small number of connected components more readily, yielding an overall trajectory that is more continuous. In contrast, ToT introduces tree-like branching and backtracking for selection, and consequently shows a richer set of $H_1$ points/bars, implying more local loops or “compare--backtrack--reselect” patterns during reasoning. Consistently, the merging process in $H_0$ is also more complex, reflecting the splitting and subsequent reintegration of semantic clusters induced by branch exploration. Furthermore, GoT displays the most pronounced increase and spread in $H_1$ features, with some points lying farther from the diagonal (i.e., higher persistence), indicating that graph-structured reasoning supports multi-path interactions and cyclic connections, thereby forming a more stable and more mesh-like topological organization. $H_0$ is more indicative of dispersion/fragmentation, whereas $H_1$ better reflects cyclicity and multi-path interactions, although an excessive amount may also suggest redundancy.

\begin{takeawaybox}
\textbf{Takeaway:} CoT is more linear with fewer loops, ToT is more complex due to branching, and GoT has the strongest multi-path connectivity with the most stable loops.
\end{takeawaybox}

\begin{figure*}
    \centering
    \includegraphics[width=0.95\linewidth]{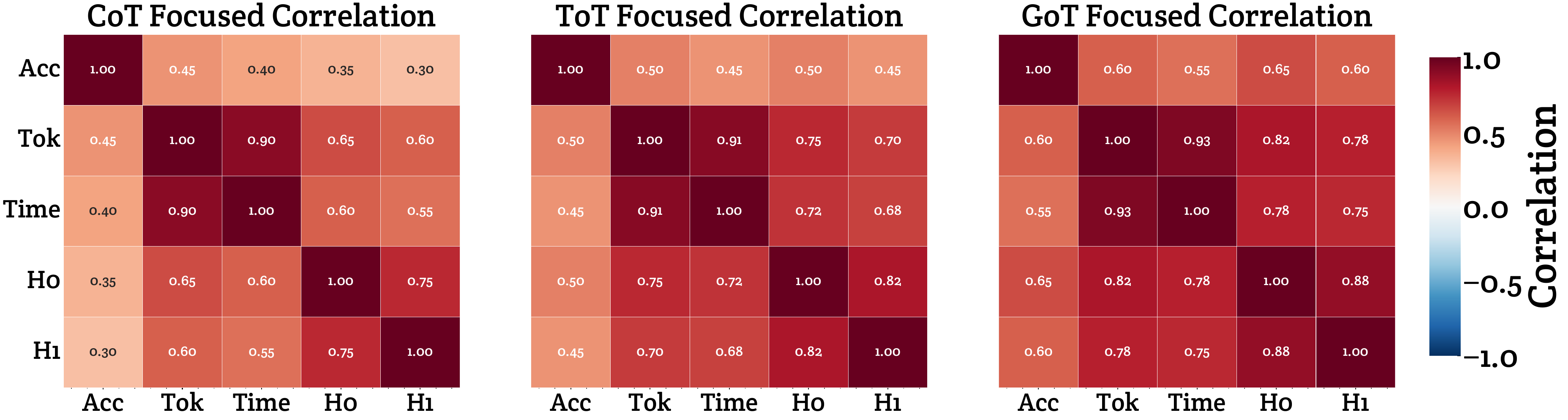}
    \caption{\textbf{Full-graph correlation heatmaps.} Three correlation heatmaps (for CoT/ToT/GoT) show the correlation structure between accuracy and metrics such as Token, Time, $H_0$, and $H_1$ under the full-graph view.}
    \label{fig:reasoning_features_heat_full}
\end{figure*}

\noindent\textbf{Feature Heatmap.} From Fig.~\ref{fig:reasoning_features_heat_full}, one can directly observe that, across the three methods, the accuracy (Acc) is positively correlated with Token, Time, $H_0$, and $H_1$. Moreover, this positive association is strongest for GoT, followed by ToT, with CoT being the weakest. This suggests that, under the full-graph view, broader exploration (more tokens/time and richer structures with more cycles) tends to coincide with higher correctness---especially for GoT, whose reasoning paradigm explicitly allows multi-path interactions and cyclic structures. Meanwhile, Tok and Time are highly correlated for all three methods, indicating that ``running longer'' essentially amounts to ``writing more.'' In addition, $H_0$ and $H_1$ are also strongly correlated, implying that fragmentation ($H_0$) and redundancy ($H_1$) often increase together.
A ``sign flip'' appears in \S\ref{sec:reasoning_features_heat_final_path}: Acc becomes negatively correlated with Tok, Time, $H_0$, and $H_1$. When focusing only on the final successful path, higher accuracy tends to arise from reasoning that is shorter circuitous, and carries less structural burden; in contrast, longer traces with more cycles look more like ``detours'' or ``self-entanglement.'' This corresponds to the phenomenon observed in strategies such as Least-to-Most, where taking too many steps can actually hurt performance and lead to lower accuracy~\cite{zhou2022least,fu2022complexity}.

\noindent\textbf{Statistical Analysis.} The statistical analysis of $H_0$ and $H_1$ (Details see \S\ref{sec:h0_statistical_analysis}) shows that lower persistent $H_0$ counts correspond to a more topologically compact point cloud with more tightly clustered word embeddings, which is associated with more coherent and smoother reasoning; in contrast, higher $H_0$ counts indicate fragmented semantics, looser conceptual connections, and poorer reasoning quality. Meanwhile, the statistics of $H_1$ suggest that effective reasoning relies, on the one hand, on the rich loop structures formed during the full exploration phase, but on the other hand ultimately converges to a more simplified topological form along the final path; long-persisting $H_1$ loops themselves are not necessarily semantically meaningful, further underscoring the importance of balancing broad exploration with focused integration.

\begin{takeawaybox}
\textbf{Takeaway:} Good reasoning = broad exploration early on (allowing loops), followed by focused integration later (collapsing the loops).
\end{takeawaybox}

\begin{figure}
    \centering
    \includegraphics[width=\linewidth]{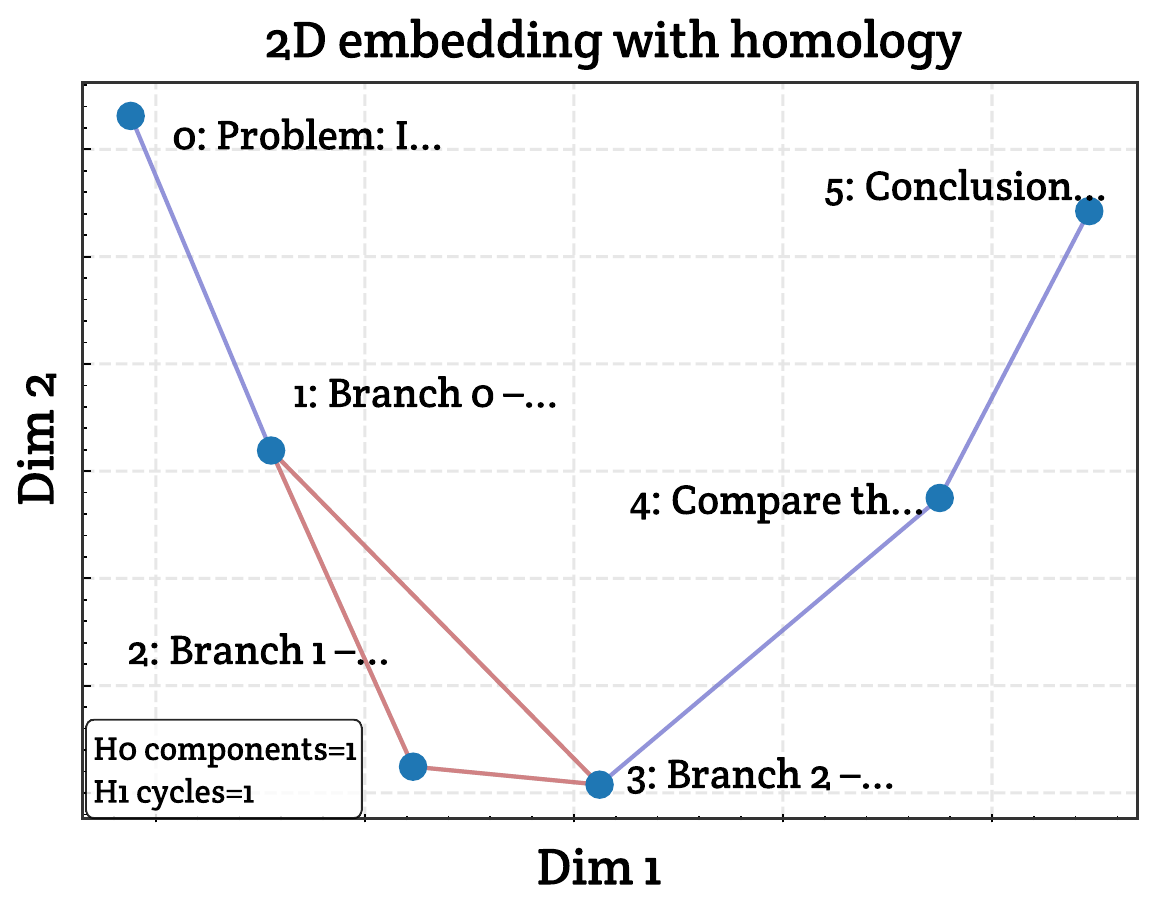}
    \caption{\textbf{Topological 2D visualization.} It visualizes a reasoning process via a 2D embedding and annotates the numbers of $H_0$ components and $H_1$ cycles.}
    \label{fig:tda_pca}
\end{figure}

\noindent\textbf{2D Visualization of Homological Structure.}
\label{sec:2D_visualization_of_homological_structure}
In this section, we take a small reasoning task---``weekend schedule planning''---as an example and perform a two-dimensional embedding and homological analysis of the reasoning process. The original data consist of six semantically clear natural-language steps (Details see \S\ref{sec:cot_weekend_planning_vis}.).
From the perspective of cognitive structure, $H_0 = 1$ indicates that the entire reasoning process consistently revolves around a single core problem, with no isolated subgraphs that deviate from the main task. Meanwhile, $H_1 = 1$ reflects the presence of nonlinear ``back-and-forth comparison'' and ``plan integration'' patterns in the reasoning: the cycle can be interpreted as the repeated trade-offs among the three pure options---study, exercise, and relaxation. These results show that, even in an extremely simple setting, homological invariants can capture comparison and integration structures that go beyond a purely linear progression of thought.
\section{Conclusion}
This work first analyzes reasoning chain quality from a structural perspective using TDA and provides a new understanding of the association between structure and reasoning ability. This work reveals that broader exploration and greater structural complexity are positively correlated with correct solutions. 
In the full-graph view, GoT enables diverse multi-path interactions and loops; however, loop-related metrics (e.g., $H_1$) alone are less predictive of accuracy than broader exploration indicators (e.g., tokens/time).
In contrast, successful reasoning outcomes in final path analysis typically feature simpler topologies, reducing redundancy and cycles, and improving interpretability and efficiency. 
This highlights the importance of balancing exploration with focused reasoning to achieve optimal outcomes. 

\section{Limitations}
Under our experimental settings (reasoning-chain length, embedding choice, and the Vietoris--Rips scale parameter $\epsilon$), we almost never observe stable higher-dimensional homology (i.e., $H_2$ and above). The few higher-dimensional features that appear typically have very short lifetimes and lie close to the diagonal, so they can be treated as noise. In CoT, ToT, and GoT, higher $H_0$ and $H_1$ loop indicators strongly correlate with a more complex topological structure, which is significantly associated with higher accuracy. 

\bibliography{custom}

@inproceedings{chiang2007discover,
title = {Discover the Semantic Topology in High-Dimensional Data},
author = {Chiang, I-Jen},
booktitle = {Expert Syst. Appl.},
year = {2007}
}

@inproceedings{wagner2012computational,
title = {Computational Topology in Text Mining},
author = {Wagner, Hubert and D{\l}otko, Pawe{\l} and Mrozek, Marian},
booktitle = {CTIC},
year = {2012}
}

@inproceedings{karlgren2014semantic,
title = {Semantic Topology},
author = {Karlgren, Jussi and Bohman, Martin and Ekgren, Ariel and Isheden, Gabriel and Kullmann, Emelie and Nilsson, David},
booktitle = {CIKM},
year = {2014}
}

@inproceedings{rathore2023topobert,
title = {{TopoBERT}: Exploring the Topology of Fine-Tuned Word Representations},
author = {Rathore, Archit and Zhou, Yichu and Srikumar, Vivek and Wang, Bei},
booktitle = {Inf. Vis.},
year = {2023}
}

@inproceedings{fitz2022shape,
title = {The Shape of Words-Topological Structure in Natural Language Data},
author = {Fitz, Stephen},
booktitle = {TAG-ML Workshops},
year = {2022}
}

@inproceedings{port2018persistent,
title = {Persistent Topology of Syntax},
author = {Port, Alexander and Gheorghita, Iulia and Guth, Daniel and Clark, John M and Liang, Crystal and Dasu, Shival and Marcolli, Matilde},
booktitle = {Math. Comput. Sci.},
year = {2018}
}

@inproceedings{savle2019topological,
title = {Topological Data Analysis for Discourse Semantics?},
author = {Savle, Ketki and Zadrozny, Wlodek and Lee, Minwoo},
booktitle = {IWCS},
year = {2019}
}

@inproceedings{carlsson2009topology,
title = {Topology and Data},
author = {Carlsson, Gunnar},
booktitle = {Bull. Amer. Math. Soc.},
year = {2009}
}

@inproceedings{zomorodian2004computing,
title = {Computing Persistent Homology},
author = {Zomorodian, Afra and Carlsson, Gunnar},
booktitle = {SoCG},
year = {2004}
}

@inproceedings{singh2007topological,
title = {Topological Methods for the Analysis of High Dimensional Data Sets and {3D} Object Recognition.},
author = {Singh, Gurjeet and M{'e}moli, Facundo and Carlsson, Gunnar E and others},
booktitle = {PBG@Eurographics},
year = {2007}
}

@inproceedings{wasserman2018topological,
title = {Topological Data Analysis},
author = {Wasserman, Larry},
booktitle = {Annu. Rev. Stat. Appl.},
year = {2018}
}

@inproceedings{turkes2022effectiveness,
title = {On the Effectiveness of Persistent Homology},
author = {Turkes, Renata and Montufar, Guido F and Otter, Nina},
booktitle = {NeurIPS},
year = {2022}
}

@inproceedings{chu2023navigate,
title = {Navigate Through Enigmatic Labyrinth a Survey of Chain of Thought Reasoning: Advances, Frontiers and Future},
author = {Chu, Zheng and Chen, Jingchang and Chen, Qianglong and Yu, Weijiang and He, Tao and Wang, Haotian and Peng, Weihua and Liu, Ming and Qin, Bing and Liu, Ting},
booktitle = {arXiv preprint arXiv:2309.15402},
year = {2023}
}

@inproceedings{yu2023towards,
title = {Towards Better Chain-of-Thought Prompting Strategies: A Survey},
author = {Yu, Zihan and He, Liang and Wu, Zhen and Dai, Xinyu and Chen, Jiajun},
booktitle = {arXiv preprint arXiv:2310.04959},
year = {2023}
}

@inproceedings{wei2022chain,
title = {Chain-of-Thought Prompting Elicits Reasoning in Large Language Models},
author = {Wei, Jason and Wang, Xuezhi and Schuurmans, Dale and Bosma, Maarten and Xia, Fei and Chi, Ed and Le, Quoc V and Zhou, Denny and others},
booktitle = {NeurIPS},
year = {2022}
}

@inproceedings{wang2022self,
title = {Self-Consistency Improves Chain of Thought Reasoning in Language Models},
author = {Wang, Xuezhi and Wei, Jason and Schuurmans, Dale and Le, Quoc and Chi, Ed and Narang, Sharan and Chowdhery, Aakanksha and Zhou, Denny},
booktitle = {arXiv preprint arXiv:2203.11171},
year = {2022}
}

@inproceedings{lu2022dynamic,
title = {Dynamic Prompt Learning via Policy Gradient for Semi-Structured Mathematical Reasoning},
author = {Lu, Pan and Qiu, Liang and Chang, Kai-Wei and Wu, Ying Nian and Zhu, Song-Chun and Rajpurohit, Tanmay and Clark, Peter and Kalyan, Ashwin},
booktitle = {arXiv preprint arXiv:2209.14610},
year = {2022}
}

@inproceedings{yao2023tree,
title = {Tree of Thoughts: Deliberate Problem Solving with Large Language Models},
author = {Yao, Shunyu and Yu, Dian and Zhao, Jeffrey and Shafran, Izhak and Griffiths, Tom and Cao, Yuan and Narasimhan, Karthik},
booktitle = {NeurIPS},
year = {2023}
}

@inproceedings{sel2023algorithm,
title = {Algorithm of Thoughts: Enhancing Exploration of Ideas in Large Language Models},
author = {Sel, Bilgehan and Al-Tawaha, Ahmad and Khattar, Vanshaj and Jia, Ruoxi and Jin, Ming},
booktitle = {arXiv preprint arXiv:2308.10379},
year = {2023}
}

@inproceedings{wang2023plan,
title = {Plan-and-Solve Prompting: Improving Zero-Shot Chain-of-Thought Reasoning by Large Language Models},
author = {Wang, Lei and Xu, Wanyu and Lan, Yihuai and Hu, Zhiqiang and Lan, Yunshi and Lee, Roy Ka-Wei and Lim, Ee-Peng},
booktitle = {arXiv preprint arXiv:2305.04091},
year = {2023}
}

@inproceedings{zhao2024kg,
title = {{KG}-{CoT}: Chain-of-Thought Prompting of Large Language Models over Knowledge Graphs for Knowledge-Aware Question Answering},
author = {Zhao, Ruilin and Zhao, Feng and Wang, Long and Wang, Xianzhi and Xu, Guandong},
booktitle = {IJCAI},
year = {2024}
}

@inproceedings{jiang2025mme,
title = {{MME}-{CoT}: Benchmarking Chain-of-Thought in Large Multimodal Models for Reasoning Quality, Robustness, and Efficiency},
author = {Jiang, Dongzhi and Zhang, Renrui and Guo, Ziyu and Li, Yanwei and Qi, Yu and Chen, Xinyan and Wang, Liuhui and Jin, Jianhan and Guo, Claire and Yan, Shen and others},
booktitle = {arXiv preprint arXiv:2502.09621},
year = {2025}
}

@inproceedings{gao2025benchmarking,
title = {Benchmarking Multimodal {CoT} Reward Model Stepwise by Visual Program},
author = {Gao, Minghe and Liu, Xuqi and Yue, Zhongqi and Wu, Yang and Chen, Shuang and Li, Juncheng and Tang, Siliang and Wu, Fei and Chua, Tat-Seng and Zhuang, Yueting},
booktitle = {arXiv preprint arXiv:2504.06606},
year = {2025}
}

@inproceedings{wang2025beyond,
title = {Beyond In-Distribution Success: Scaling Curves of {CoT} Granularity for Language Model Generalization},
author = {Wang, Ru and Huang, Wei and Song, Selena and Zhang, Haoyu and Iwasawa, Yusuke and Matsuo, Yutaka and Guo, Jiaxian},
booktitle = {arXiv preprint arXiv:2502.18273},
year = {2025}
}

@inproceedings{zhou2025landscape,
title = {Landscape of Thoughts: Visualizing the Reasoning Process of Large Language Models},
author = {Zhou, Zhanke and Zhu, Zhaocheng and Li, Xuan and Galkin, Mikhail and Feng, Xiao and Koyejo, Sanmi and Tang, Jian and Han, Bo},
booktitle = {arXiv preprint arXiv:2503.22165},
year = {2025}
}

@inproceedings{munch2017user,
title = {A User’s Guide to Topological Data Analysis},
author = {Munch, Elizabeth},
booktitle = {J. Learn. Anal.},
year = {2017}
}

@inproceedings{chazal2021introduction,
title = {An Introduction to Topological Data Analysis: Fundamental and Practical Aspects for Data Scientists},
author = {Chazal, Fr{'e}d{'e}ric and Michel, Bertrand},
booktitle = {Front. Artif. Intell.},
year = {2021}
}

@inproceedings{zhou2022learning,
title = {Learning Persistent Homology of {3D} Point Clouds},
author = {Zhou, Chi and Dong, Zhetong and Lin, Hongwei},
booktitle = {Comput. Graph.},
year = {2022}
}

@inproceedings{nils2019sentence,
title = {Sentence-{BERT}: Sentence Embeddings Using {S}iamese {BERT}-Networks},
author = {Reimers, Nils and Gurevych, Iryna},
booktitle = {EMNLP-IJCNLP},
year = {2019}
}

@inproceedings{besta2024graph,
title = {Graph of Thoughts: Solving Elaborate Problems with Large Language Models},
author = {Besta, Maciej and Blach, Nils and Kubicek, Ales and Gerstenberger, Robert and Podstawski, Michal and Gianinazzi, Lukas and Gajda, Joanna and Lehmann, Tomasz and Niewiadomski, Hubert and Nyczyk, Piotr and others},
booktitle = {AAAI},
year = {2024}
}

@inproceedings{zhou2022least,
title = {Least-to-Most Prompting Enables Complex Reasoning in Large Language Models},
author = {Zhou, Denny and Sch{"a}rli, Nathanael and Hou, Le and Wei, Jason and Scales, Nathan and Wang, Xuezhi and Schuurmans, Dale and Cui, Claire and Bousquet, Olivier and Le, Quoc and others},
booktitle = {arXiv preprint arXiv:2205.10625},
year = {2022}
}

@inproceedings{ning2023skeleton,
title = {Skeleton-of-Thought: Large Language Models Can Do Parallel Decoding},
author = {Ning, Xuefei and Lin, Zinan and Zhou, Zixuan and Wang, Zifu and Yang, Huazhong and Wang, Yu},
booktitle = {ENLSP-III},
year = {2023}
}

@inproceedings{madaan2023self,
title = {Self-Refine: Iterative Refinement with Self-Feedback},
author = {Madaan, Aman and Tandon, Niket and Gupta, Prakhar and Hallinan, Skyler and Gao, Luyu and Wiegreffe, Sarah and Alon, Uri and Dziri, Nouha and Prabhumoye, Shrimai and Yang, Yiming and others},
booktitle = {NeurIPS},
year = {2023}
}

@inproceedings{zheng2023take,
title = {Take a Step Back: Evoking Reasoning via Abstraction in Large Language Models},
author = {Zheng, Huaixiu Steven and Mishra, Swaroop and Chen, Xinyun and Cheng, Heng-Tze and Chi, Ed H and Le, Quoc V and Zhou, Denny},
booktitle = {arXiv preprint arXiv:2310.06117},
year = {2023}
}

@inproceedings{hendrycksmath2021,
title = {Measuring Mathematical Problem Solving With the {MATH} Dataset},
author = {Dan Hendrycks and Collin Burns and Saurav Kadavath and Akul Arora and Steven Basart and Eric Tang and Dawn Song and Jacob Steinhardt},
booktitle = {NeurIPS},
year = {2021}
}

@inproceedings{sun2025survey,
title = {A Survey of Reasoning with Foundation Models: Concepts, Methodologies, and Outlook},
author = {Sun, Jiankai and Zheng, Chuanyang and Xie, Enze and Liu, Zhengying and Chu, Ruihang and Qiu, Jianing and Xu, Jiaqi and Ding, Mingyu and Li, Hongyang and Geng, Mengzhe and others},
booktitle = {ACM Comput. Surv.},
year = {2025}
}

@inproceedings{aggarwal2025l1,
title = {{L1}: Controlling How Long a Reasoning Model Thinks with Reinforcement Learning},
author = {Aggarwal, Pranjal and Welleck, Sean},
booktitle = {arXiv preprint arXiv:2503.04697},
year = {2025}
}

@inproceedings{luo2025deconstructing,
title = {Deconstructing Long Chain-of-Thought: A Structured Reasoning Optimization Framework for Long {CoT} Distillation},
author = {Luo, Yijia and Song, Yulin and Zhang, Xingyao and Liu, Jiaheng and Wang, Weixun and Chen, GengRu and Su, Wenbo and Zheng, Bo},
booktitle = {arXiv preprint arXiv:2503.16385},
year = {2025}
}

@inproceedings{he2025can,
title = {Can Large Language Models Detect Errors in Long Chain-of-Thought Reasoning?},
author = {He, Yancheng and Li, Shilong and Liu, Jiaheng and Wang, Weixun and Bu, Xingyuan and Zhang, Ge and Peng, Zy and Zhang, Zhaoxiang and Zheng, Zhicheng and Su, Wenbo and others},
booktitle = {ACL},
year = {2025}
}

@inproceedings{yeo2025demystifying,
title = {Demystifying Long Chain-of-Thought Reasoning in {LLM}s},
author = {Yeo, Edward and Tong, Yuxuan and Niu, Morry and Neubig, Graham and Yue, Xiang},
booktitle = {arXiv preprint arXiv:2502.03373},
year = {2025}
}

@inproceedings{feng2025efficient,
title = {Efficient Reasoning Models: A Survey},
author = {Feng, Sicheng and Fang, Gongfan and Ma, Xinyin and Wang, Xinchao},
booktitle = {arXiv preprint arXiv:2504.10903},
year = {2025}
}

@inproceedings{liu2025logic,
title = {Logic-of-Thought: Injecting Logic into Contexts for Full Reasoning in Large Language Models},
author = {Liu, Tongxuan and Xu, Wenjiang and Huang, Weizhe and Zeng, Yuting and Wang, Jiaxing and Wang, Xingyu and Yang, Hailong and Li, Jing},
booktitle = {NAACL},
year = {2025}
}

@inproceedings{jin2025zero,
title = {Zero-Shot Chain-of-Thought Reasoning Guided by Evolutionary Algorithms in Large Language Models},
author = {Jin, Feihu and Liu, Yifan and Tan, Ying},
booktitle = {NLPCC},
year = {2025}
}

@inproceedings{zhang2025improve,
title = {Improve Vision Language Model Chain-of-Thought Reasoning},
author = {Zhang, Ruohong and Zhang, Bowen and Li, Yanghao and Zhang, Haotian and Sun, Zhiqing and Gan, Zhe and Yang, Yinfei and Pang, Ruoming and Yang, Yiming},
booktitle = {ACL},
year = {2025}
}

@inproceedings{qu2025survey,
title = {A Survey of Efficient Reasoning for Large Reasoning Models: Language, Multimodality, and Beyond},
author = {Qu, Xiaoye and Li, Yafu and Su, Zhaochen and Sun, Weigao and Yan, Jianhao and Liu, Dongrui and Cui, Ganqu and Liu, Daizong and Liang, Shuxian and He, Junxian and others},
booktitle = {arXiv preprint arXiv:2503.21614},
year = {2025}
}

@inproceedings{cobbe2021gsm8k,
  title={Training Verifiers to Solve Math Word Problems},
  author={Cobbe, Karl and Kosaraju, Vineet and Bavarian, Mohammad and Chen, Mark and Jun, Heewoo and Kaiser, Lukasz and Plappert, Matthias and Tworek, Jerry and Hilton, Jacob and Nakano, Reiichiro and Hesse, Christopher and Schulman, John},
  booktitle={arXiv preprint arXiv:2110.14168},
  year={2021}
}

@inproceedings{hendryckstest2021,
  title={Measuring Massive Multitask Language Understanding},
  author={Dan Hendrycks and Collin Burns and Steven Basart and Andy Zou and Mantas Mazeika and Dawn Song and Jacob Steinhardt},
  booktitle={ICLR},
  year={2021}
}

@inproceedings{fu2022complexity,
  title={Complexity-based prompting for multi-step reasoning},
  author={Fu, Yao and Peng, Hao and Sabharwal, Ashish and Clark, Peter and Khot, Tushar},
  booktitle={ICLR},
  year={2022}
}
\newpage
\appendix

\section{Appendix}
\label{sec:appendix}

\subsection{Preliminaries}

\noindent\textbf{Vietoris–Rips Complex.}
\label{sec:pre-vietoris_rips_complex}
The Vietoris–Rips complex is a fundamental construction in computational topology for extracting the shape of a point cloud at a given scale. 
Let \( (X, d) \) be a finite metric space, where \(X\) is a set of points and \(d\) is a distance function on \(X\). 
For a scale parameter \(\epsilon > 0\), the Vietoris–Rips complex \(\mathrm{VR}_\epsilon(X)\) is the abstract simplicial complex whose vertex set is \(X\), and where a finite subset \(\sigma \subseteq X\) forms a simplex if and only if all pairwise distances between points in \(\sigma\) are at most \(\epsilon\):
\begin{equation}
    \mathrm{VR}_\epsilon(X)
    = \bigl\{ \sigma \subseteq X \,\big|\,
    d(x_i, x_j) \le \epsilon,\ \forall x_i, x_j \in \sigma \bigr\}.
\end{equation}
Intuitively, \(\mathrm{VR}_\epsilon(X)\) connects points that are \(\epsilon\)-close and fills in higher-dimensional simplices whenever a group of points is mutually close. 
By varying the scale \(\epsilon\), one obtains a nested family of complexes that captures multi-scale topological features (such as connected components and loops) of the underlying data and serves as the basis for persistent homology.

\noindent\textbf{Topological Features.}
\label{sec:topological_features}
In persistent homology analysis, 0-cycles and 1-cycles are two important topological features, representing connectivity and circular structures at different scales. 

\noindent\textbf{0-cycle.} The 0-cycle represents the connected components in the data. At a given scale, points connected together form a connected branch, also called a "0-cycle." As the scale of the point cloud increases, these connected branches may merge, disappear, or continue to expand. 

First, we define a radius sequence \( r_0 > r_1 > \cdots > r_m \) and construct the corresponding simplicial complex \( \mathrm{VR}_{r_i}(X) \). For each simplicial complex \( \mathrm{VR}_{r_i}(X) \), the rank of its 0-dimensional homology group \( H_0(\mathrm{VR}_{r_i}(X)) \) represents the number of connected branches in the complex. That is, the rank \( \text{rank}(H_0(\mathrm{VR}_{r_i}(X))) \) reflects the number of distinct connected components or branches in the complex. When the radius \( r \) decreases to a stable radius \( r_s \), the number of connected branches no longer changes. The stable number of connected branches is denoted as \( n_i \):
\begin{equation}
    \exists r_i > 0, \forall r < r_i, \text{rank}(H_0(\mathrm{VR}_{r}(X))) = n_i
\end{equation}

For the point cloud \( X \) corresponding to a reasoning chain, the change in its 0-cycle reflects the tightness of the reasoning logic. When the coordinates in the word embedding space are closer, the logical flow of the reasoning chain is tighter, meaning stronger semantic coherence. Conversely, if there are more connected branches, it indicates more disconnection or lack of relations between steps in the reasoning process, leading to weaker coherence.

\noindent\textbf{1-cycle.} A 1-cycle represents circular structures, reflecting redundant logical paths in the reasoning chain. Elements in \( H_1(\mathrm{VR}_{r_i}(X)) \) represent the 1-cycle at that scale, showing potential redundant loops or repetitive paths in the reasoning process. From a topological viewpoint, if a 1-cycle persists over a wide scale interval \( [b, d) \), it indicates stability in the topological structure, reflecting the ongoing presence of redundant paths in the reasoning chain. Semantically, a persistent 1-cycle may correspond to a "formal logical closure" path in reasoning. These paths may follow formal logical rules but lack meaningful content, possibly leading to logically sound but absurd conclusions. For example, seemingly valid reasoning paths may conform to local logic but ultimately result in false or meaningless outcomes. Thus, a long-lasting 1-cycle does not necessarily represent a meaningful semantic structure. In the word embedding space, such structures may simply be closed arrangements of local vectors, not equivalent to valid reasoning. In other words, persistent redundant paths may not contribute to actual reasoning progress but are just local patterns in the data.

\subsection{Reasoning Step Embedding}
\label{subsec:semantic-embedding}

This section provides a detailed account of how chain-structured (CoT), tree-structured (ToT), and graph-structured (GoT) reasoning processes are mapped into point-cloud representations that preserve both geometric consistency and structural sensitivity. We further discuss the motivation, structural constraints, and theoretical underpinnings of the positional encoding designs used for each type of reasoning structure.

\noindent\textbf{Motivation: Why Embedding Matters for Topological Analysis.}
TDA focuses on characterizing connectivity, cycles, and higher-dimensional voids across multiple scales. Its analytical outcomes are highly sensitive to the geometric structure of the input point cloud. Therefore, the embedding of reasoning steps must satisfy the following requirements: (1) Semantically similar reasoning steps should be mapped to nearby points in the space, ensuring that the resulting topological structure reflects semantic continuity rather than random noise. (2) Structural relationships among reasoning steps--such as sequential order, hierarchical depth, or graph connectivity—must be preserved in the geometry of the point cloud, otherwise persistent homology would be unable to distinguish between different types of reasoning structures. (3) Embeddings for different classes of reasoning structures must maintain structural consistency, so that topological indicators derived from CoT, ToT, and GoT are directly comparable. (4) Local geometric patterns should capture neighborhood relationships between reasoning steps, while global geometry should represent the overall organization of the reasoning chain, aligning with TDA’s requirement to analyze data across multiple scales.

Based on these considerations, we integrate semantic embedding with structure-aware positional encoding to ensure that the embedded reasoning chains exhibit both semantic continuity and structurally interpretable geometric patterns in high-dimensional space.

\noindent\textbf{Semantic Embedding of Reasoning Steps.}
For a reasoning chain
\begin{align}
    S = (s_1, s_2, \dots, s_n),
\end{align}
each reasoning step \(s_i\) is a natural language sentence that may contain logical transitions, mathematical computations, intermediate inferences, hypothesis updates, and other semantic content. To represent these steps in a high-dimensional space, we adopt a semantic encoding function
\begin{align}
    \Phi : L \to \mathbb{R}^d
\end{align}
and obtain
\begin{align}
    \vec{x}_i = \Phi(s_i),
\end{align}
where \(L\) denotes the space of all possible natural language sentences.

In practice, semantic embeddings are typically derived from pretrained language models such as BERT, GPT, or sentence-transformer models. In this work, we specifically use the all-mpnet-base-v2 model from the sentence-transformers~\citep{nils2019sentence} library as our sentence encoder. This choice offers several advantages: (1) It captures sentence-level semantic consistency, allowing semantically similar reasoning steps to be mapped to nearby regions in the embedding space. (2) It preserves smooth semantic variation along the reasoning chain, reflecting gradual evolution of the model's internal reasoning state. (3) It leverages the structured semantic representations already learned in the latent space of large pretrained language models.

However, semantic embeddings alone primarily encode ``content similarity'' and do not explicitly model structural properties of the reasoning process, such as sequential order, branching patterns, or graph connectivity. To address this limitation, we augment the semantic embeddings with structure-aware positional encodings, which we describe in the subsequent subsections.

\noindent\textbf{Chain-of-Thought: Sequential Positional Encoding.}
CoT treats the reasoning process as a linearly progressing sequence:
\begin{align}
    s_1 \rightarrow s_2 \rightarrow \dots \rightarrow s_n.
\end{align}
This structure has two key characteristics: (1) strict order sensitivity; (2) single-path progress.

To model this structure, we adopt the classical sinusoidal positional encoding used in Transformers:

\begin{align}
    \text{PE}_k(i) =
    \begin{cases}
        \sin\!\left( \dfrac{i}{10000^{k/d}} \right),           & \text{if } k \text{ is even}, \\[6pt]
        \cos\!\left( \dfrac{i}{10000^{(k-1)/d}} \right),      & \text{if } k \text{ is odd}.
    \end{cases}
\end{align}

The final embedding is then given by:
\begin{align}
    \widetilde{\vec{x}_i} = \vec{x}_i + \text{PE}(i).
\end{align}

\begin{takeawaybox}
Motivation: Multi-frequency encoding essentially maps the CoT onto a low-dimensional curve in a high-dimensional space that is locally continuous, globally separable, and preserves \(H_0\) connectivity.
\end{takeawaybox}

Multi-frequency encoding assigns different periods to different dimensions. This design allows the model to distinguish between different positional relationships in the embedding space, since each dimension responds to positional changes with its own characteristic frequency.
The resulting representation is locally continuous yet globally distinguishable. Neighboring reasoning steps are mapped to nearby points, while steps that are far apart in the reasoning chain still retain clear structural differences in the embedding space.
Moreover, the encoding preserves the linear structural and topological characteristics of the reasoning chain. In the point cloud of embeddings, the reasoning process tends to form a single continuous trajectory, which facilitates the capture of the evolution of \(H_0\) connectivity. Consequently, the embedding of a CoT tends to lie along a low-dimensional manifold that appears as a curve embedded in a high-dimensional space.

\noindent\textbf{Tree-of-Thought: Depth and Branch Encoding.}
ToT extends a linear reasoning chain to a tree-structured process. Each node in the tree is assigned a depth \(d_i\), indicating its logical stage, and a branch index \(b_i\), indicating the reasoning path from which it originates. We introduce two types of structural positional encodings and combine them with the original token embedding:
\begin{align}
\widetilde{\vec{x}_i}
= \vec{x}_i
- PE_{\mathrm{depth}}(d_i)
- PE_{\mathrm{branch}}(b_i).
\end{align}

\begin{takeawaybox}
Motivation: Depth and branch encodings essentially write the tree's ``hierarchy'' and ``branching'' into the embedding, making the ToT point cloud exhibit a clearly layered, tree-like fan structure.
\end{takeawaybox}

The depth encoding emphasizes the layer-wise unfolding of logical reasoning. By assigning distinct depth-dependent offsets, it separates nodes belonging to different logical stages in the geometric space and reduces the risk of confusion across depths. Geometrically, the resulting point cloud exhibits a layered structure, and topological data analysis (TDA) can observe depth-dependent changes in \(H_0\) connectivity as one moves from shallow to deeper layers.
The branch encoding differentiates distinct reasoning paths that may occur at the same depth. It prevents nodes at the same layer from collapsing into a single semantic cluster and instead spreads them into multiple, relatively separated clusters in the embedding space. From a topological perspective, this increased diversity of local paths can lead to more short-lived \(H_1\) features in persistent homology, which correspond to exploratory loops in the reasoning structure.
Both encodings are indispensable because the tree structure is inherently hierarchical and branching. Encoding only depth would collapse different branches at the same level and effectively erase the branching structure. Encoding only branches would ignore the hierarchy and fail to preserve the layered organization of reasoning. With both depth and branch encodings combined, the ToT point cloud typically exhibits a tree-like, fan-shaped distribution, with clear hierarchical expansion as depth increases.

\noindent\textbf{Graph-of-Thought: Laplacian Eigenvector Encoding.}
GoT provides the richest form of structural expressiveness. Its reasoning process can be organized as an arbitrary graph, allowing for multiple parallel paths, cyclic reasoning, path merging, cross-path interactions, and repeated exploration of local regions. Unlike chains or trees, such a graph does not possess a fixed global order or a simple hierarchical structure.

To encode this general graph structure, we adopt a Laplacian eigenvector-based positional encoding. Let the graph Laplacian be
\begin{align}
L = D - A,
\end{align}
where \(D\) is the degree matrix and \(A\) is the adjacency matrix. We perform the eigen-decomposition
\begin{align}
L U = \Lambda U,
\end{align}
where the rows \(U_i \in \mathbb{R}^d\) correspond to the eigenvector-based coordinates of node \(i\). The positional encoding of node \(i\) is then defined as
\begin{align}
PE_{\mathrm{graph}}(i) = U_i,
\end{align}
and the final embedding for node \(i\) is given by
\begin{align}
\widetilde{\vec{x}_i} = \vec{x}_i + PE_{\mathrm{graph}}(i).
\end{align}

\begin{takeawaybox}
Motivation: Laplacian eigenvector encoding essentially uses the graph's spectral structure to embed nodes, so that GoT preserves global topological patterns and rich loops and cavities (\(H_1, H_2\)) in the point cloud.
\end{takeawaybox}

Laplacian eigenvectors are used because they capture the global manifold structure of the graph. Low-frequency eigenvectors reveal large-scale connectivity patterns, community boundaries, and flow-like structures across the graph. At the same time, the embedding preserves geometric consistency: adjacent nodes tend to be mapped to nearby points, whereas nodes that are not connected, or are far apart in the graph, are automatically pushed farther away in the embedding space.
This construction naturally supports cycles, holes, and higher-order topological features. Complex structures such as multiple loops and void-like regions are preserved in the point cloud, allowing TDA to detect higher-dimensional homology groups such as \(H_1\) and \(H_2\). Another advantage is that Laplacian-based embedding does not require manual specification of an explicit structural dimension; it is a standard, stable method in graph embedding.
As a result, GoT embeddings typically form a network-like connected structure in the point cloud. One can observe multiple \(H_1\) loops, locally closed regions that may give rise to \(H_2\) features, and intricate crossings between different reasoning paths. These properties align well with the requirements of TDA for representing and analyzing complex reasoning trajectories.

\subsection{Experimental Metrics and Quantification}
\label{sec:experimental_Metrics&quantification}

\begin{figure}
    \centering
    \includegraphics[width=0.9\linewidth]{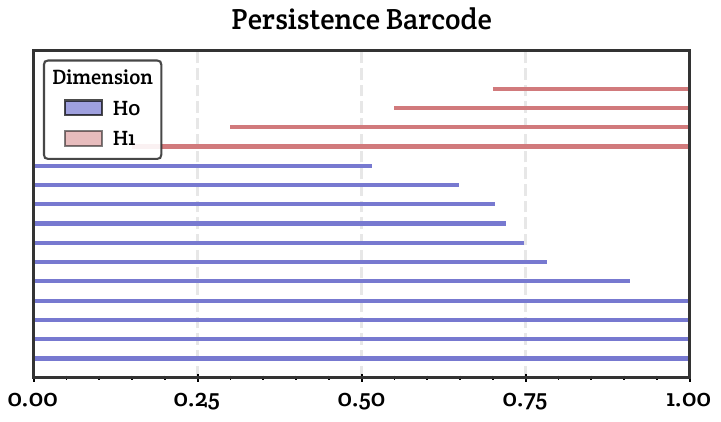}
    \caption{TDA persistence barcode.}
    \label{fig:barcode_eg}
\end{figure}
\noindent\textbf{Barcode Diagram.}
As shown in Fig.~\ref{fig:barcode_eg}, the barcode diagram is an effective tool for displaying the “lifecycle’’ of topological features (such as connected components, loops, cavities, etc.) at different scales. Each horizontal bar (barcode) corresponds to the existence interval of a topological feature, with the \(x\)-axis representing the scale (typically the parameter \(\epsilon\)), and the starting and ending points of the bar indicating the birth and death scales of the feature. Multi-dimensional barcode diagrams can plot multiple groups of bars to represent topological features in different homological dimensions, such as 0-dimensional connected components, 1-dimensional loops, and so on. The barcode diagram provides an intuitive view of the persistence of features across scales: the longer the barcode, the more stable the feature is in the data, indicating a longer duration at larger scales; conversely, shorter barcodes represent features that disappear at smaller scales, potentially indicating noise or unimportant topological structures in the data.

\begin{figure}
    \centering
    \includegraphics[width=\linewidth]{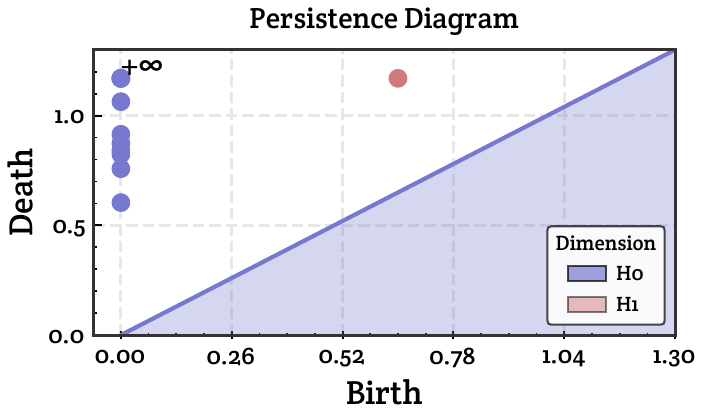}
    \caption{TDA persistence diagram.}
    \label{fig:persistence_diagram_eg}
\end{figure}
\noindent\textbf{Persistence Diagram.}
As shown in Fig.~\ref{fig:persistence_diagram_eg}, the persistence diagram is another effective visualization tool used to represent the lifecycle of topological features across scale changes. In a persistence diagram, each topological feature corresponds to a point in a 2D plane, with the \(x\)-axis representing the feature’s birth scale \(b\) and the \(y\)-axis representing its death scale \(d\). Thus, the \(x\)-axis encodes the scale at which the feature first appears, while the \(y\)-axis encodes the scale at which the feature disappears or merges. The diagonal line \(d = b\) represents the limit of a “lifetime of zero,’’ indicating a feature that disappears as soon as it is born, typically corresponding to noise in the data. Points that lie farther from the diagonal indicate features with longer lifetimes, meaning these features are more stable and significant in the data, exhibiting higher “persistence.’’

\noindent\textbf{Total Lifetime Length} refers to the overall duration of existence of all topological features across all scales:
    \begin{equation}
         L = \sum_{i=1}^{N}l_i=\sum_{i=1}^{N} (d_i - b_i)
    \end{equation}
    where \(d_i\) and \(b_i\) are the death and birth scales of the \(i\)-th feature, respectively.
\noindent\textbf{Average Lifetime} denotes the average lifespan of topological features:
    \begin{equation}
         \bar{l} = \frac{1}{N}\sum_{i=1}^{N}l_i
    \end{equation}
\noindent\textbf{Maximum Lifetime} represents the longest lifetime of any topological feature:
    \begin{equation}
            l_{max} = \max\limits_{x} l_i
    \end{equation}
\noindent\textbf{Lifetime Variance} quantifies the spread or variability of topological feature lifetimes, indicating the stability and distribution of these features across scales.

    \begin{equation}
        Var(l)  = \frac{1}{N} \sum_{i=1}^{N} (l_i - \bar{l})^2
    \end{equation}
These statistical metrics provide a comprehensive view of the topological structure's stability and the concentration of feature lifetimes across different scales.

For each homology dimension \( k \), the number of intervals in the corresponding Barcode Diagram is counted, denoted as \( n_k \), which represents the number of distinct topological features in that dimension. Specifically, \( n_k \) is calculated by:
\begin{equation}
     n_k = \text{Number of intervals in } H_k 
\end{equation}
\( n_k \) reflects the structural complexity of the topological features in the data. For instance, a larger \( n_k \) indicates more loop structures, suggesting a more complex path structure in the data.

\noindent\textbf{Persistent Entropy \(H\)} measures the distribution complexity of topological information by quantifying how evenly the lifetimes of topological features are distributed across scales. It is computed in two steps: first, the normalized lifetime weight $p_i$ for each topological feature is calculated as:
\begin{equation}
    p_i = \frac{l_i}{\sum_{j=1}^{N} l_j}
\end{equation}
where \(l_i = d_i - b_i\) represents the lifetime of the \(i\)-th feature. Then, the \(H\) is computed as:
\begin{equation}
    H = -\sum_{i}^{N} p_i \log p_i
\end{equation}
A higher \(H\) indicates that the distribution of topological feature lifetimes is more uniform, suggesting that the structure is more ``random" or dispersed across many features. Conversely, a lower \(H\) implies that there are significant differences among feature lifetimes, meaning that a few dominant structures persist much longer than others.

\subsection{Additional Experiments}
\label{sec:additional_experiments}

\begin{table*}[t]
\centering
\small
\scalebox{0.8}{
\begin{tabular}{l|c|c c c c c c c}
\toprule
Dataset & Method & Acc. & $|H_0|$ & $H_0^{\mathrm{avg}}$ &  $H_0^{\max}$ & $|H_1|$
& $H_1^{\max}$ & $H_1^{\mathrm{avg}}$ \\
\midrule
\multirow{3}{*}{GSM8K~\cite{cobbe2021gsm8k} (3.5-turbo)} 
  & CoT~\cite{wei2022chain} & 0.678 & 2.853 & 0.043 & 0.029 & 0.089 & 0.015 & 0.012 \\
  & ToT~\cite{yao2023tree} & 0.703 & 3.666 & 0.046 & 0.031 & 0.259 & 0.041 & 0.022 \\
  & GoT~\cite{besta2024graph} & 0.737 & 4.070 & 0.049 & 0.033 & 0.283 & 0.038 & 0.024 \\
\midrule
\multirow{3}{*}{MATH~\cite{hendrycksmath2021} (3.5-turbo)} 
  & CoT~\cite{wei2022chain} & 0.433 & 2.691 & 0.037 & 0.024 & 0.143 & 0.024 & 0.017 \\
  & ToT~\cite{yao2023tree} & 0.516 & 3.418 & 0.044 & 0.029 & 0.255 & 0.040 & 0.023 \\
  & GoT~\cite{besta2024graph} & 0.600 & 3.898 & 0.049 & 0.032 & 0.295 & 0.040 & 0.023 \\
\midrule
\multirow{3}{*}{MMLU~\cite{hendryckstest2021} (3.5-turbo)} 
  & CoT~\cite{wei2022chain} & 0.545 & 3.061 & 0.041 & 0.026 & 0.035 & 0.006 & 0.006 \\
  & ToT~\cite{yao2023tree} & 0.517 & 4.028 & 0.050 & 0.033 & 0.267 & 0.039 & 0.021 \\
  & GoT~\cite{besta2024graph} & 0.529 & 4.497 & 0.055 & 0.036 & 0.357 & 0.044 & 0.023 \\
\midrule
\multirow{3}{*}{GSM8K~\cite{cobbe2021gsm8k} (4o-mini)} 
  & CoT~\cite{wei2022chain} & 0.670 & 2.050 & 0.030 & 0.019 & 0.080 & 0.003 & 0.002 \\
  & ToT~\cite{yao2023tree} & 0.755 & 3.600 & 0.046 & 0.030 & 0.265 & 0.040 & 0.024 \\
  & GoT~\cite{besta2024graph} & 0.790 & 5.200 & 0.060 & 0.040 & 0.700 & 0.058 & 0.026 \\
\midrule
\multirow{3}{*}{MATH~\cite{hendrycksmath2021} (4o-mini)} 
  & CoT~\cite{wei2022chain} & 0.475 & 2.076 & 0.031 & 0.020 & 0.116 & 0.003 & 0.002 \\
  & ToT~\cite{yao2023tree} & 0.617 & 3.489 & 0.045 & 0.029 & 0.253 & 0.043 & 0.026 \\
  & GoT~\cite{besta2024graph} & 0.657 & 4.205 & 0.052 & 0.034 & 0.563 & 0.063 & 0.027 \\
\midrule
\multirow{3}{*}{MMLU~\cite{hendryckstest2021} (4o-mini)} 
  & CoT~\cite{wei2022chain} & 0.529 & 2.070 & 0.031 & 0.020 & 0.043 & 0.002 & 0.002 \\
  & ToT~\cite{yao2023tree} & 0.521 & 3.825 & 0.048 & 0.031 & 0.280 & 0.035 & 0.021 \\
  & GoT~\cite{besta2024graph} & 0.579 & 6.450 & 0.074 & 0.048 & 0.921 & 0.061 & 0.023 \\
\bottomrule
\end{tabular}
}
\caption{Performance of reasoning chains across different structural indicators.}
\label{tab:tda_all_metrics}
\end{table*}

\begin{figure*}
    \centering
    \includegraphics[width=0.95\linewidth]{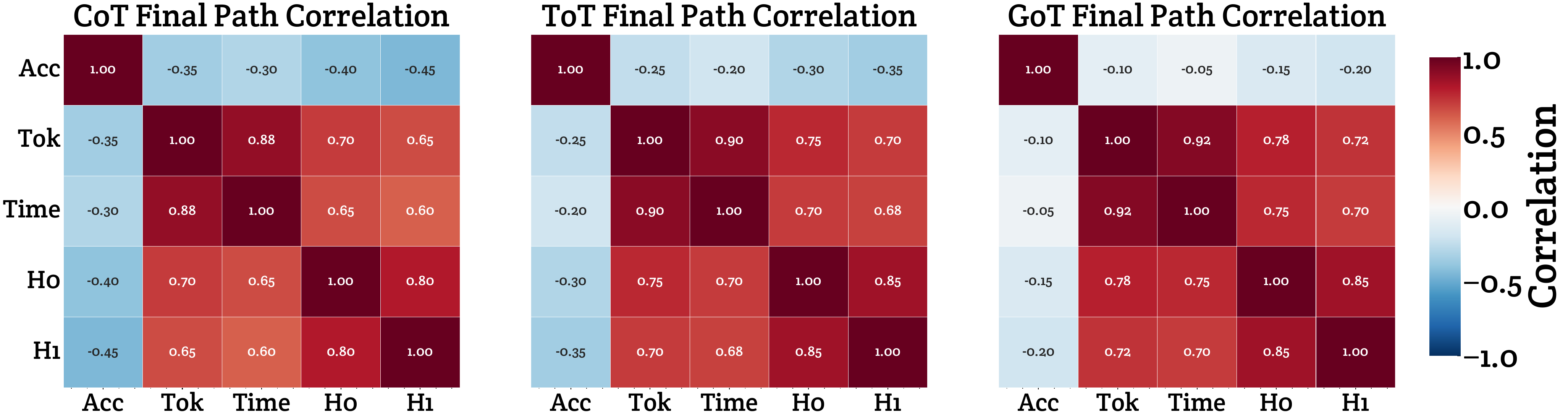}
    \caption{Final-path correlation heatmaps of topological features and performance for CoT, ToT, and GoT.}
    \label{fig:reasoning_features_heat_final}
\end{figure*}

\noindent\textbf{Final Path Analysis.} 
\label{sec:reasoning_features_heat_final_path}
The bottom portion of Fig.~\ref{fig:reasoning_features_heat_final} presents the final path correlation heatmap, where successful outcomes are characterized by logical clarity and directness, reflected in simpler topological structures. In these cases, the convergent paths exhibit streamlined correlations with minimal redundancy or cycles, indicating that insights from the exploration phase have been focused and refined. The simplicity of this topological structure ensures the interpretability and efficiency of the final reasoning, aligning with verifiable logic and reducing cognitive overhead and potential errors in the validation process.

\begin{figure*}
    \centering
    \includegraphics[width=\linewidth]{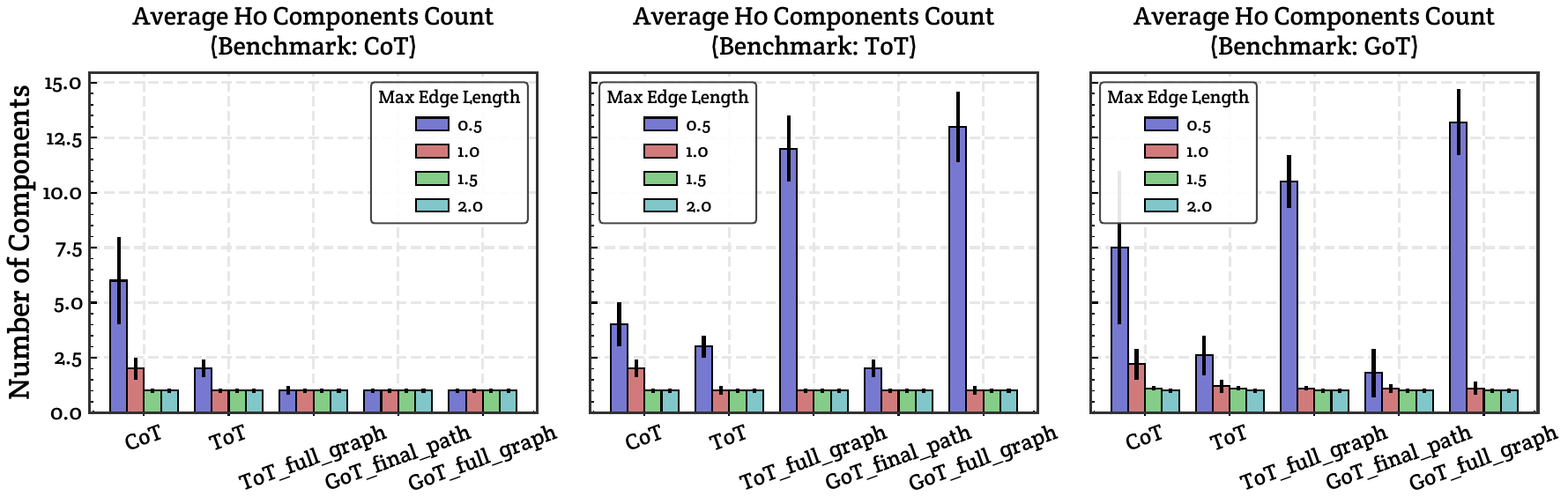}
    \caption{The number of $H_0$ components.}
    \label{$H_0$nmuber}
\end{figure*}

\noindent\textbf{Statistical Analysis Setting.} The dataset is constructed following a ``benchmark-paradigm sampling + cross-paradigm re-run'' protocol. Specifically, we use CoT, ToT, and GoT as benchmark reasoning paradigms, respectively, and first collect a fixed number of correctly and incorrectly answered problems from the MATH dataset to form three small-scale subsets. Then, for each subset, we feed the exact same set of problems to all three paradigms (CoT/ToT/GoT) to obtain comparable reasoning outcomes and corresponding thought structures.
The experimental procedure is identical across the three benchmark subsets. For each sample, we first save the thought structure generated during reasoning. We then perform TDA on the saved structure: under different values of \( \epsilon \) (0.5, 1.0, 1.5, 2.0), we construct a filtration process and compute persistent-homology artifacts and metrics, which are used to characterize the topological differences among reasoning paradigms in terms of both global exploration (full graph) and final-path convergence.

\noindent\textbf{$H_0$ Statistical Analysis.} 
\label{sec:h0_statistical_analysis}
We perform a statistical analysis of the components in each homology group and present the results as bar charts. Fig.~\ref{$H_0$nmuber} shows that as the filtration scale increases, the number of connected components in $H_0$ decreases until stabilizing, representing the number of permanently persistent components. For the point cloud corresponding to the reasoning chain, this stable value $a$ is inversely related to the logical coherence of the reasoning: smaller values of $a$ indicate closer distances between word embeddings, promoting tighter logical connections. This results in a more unified topological structure, where semantic elements merge early in the filtration process, reducing fragmentation and improving the smoothness of reasoning. In contrast, larger $a$ values suggest scattered clusters, possibly reflecting loosely connected ideas, which hinder convergence to a solution. These findings suggest that topological compactness, indicated by low persistent $H_0$ counts, can serve as a quantifiable metric for evaluating the reasoning quality in language models.

\begin{figure*}
    \centering
    \includegraphics[width=\linewidth]{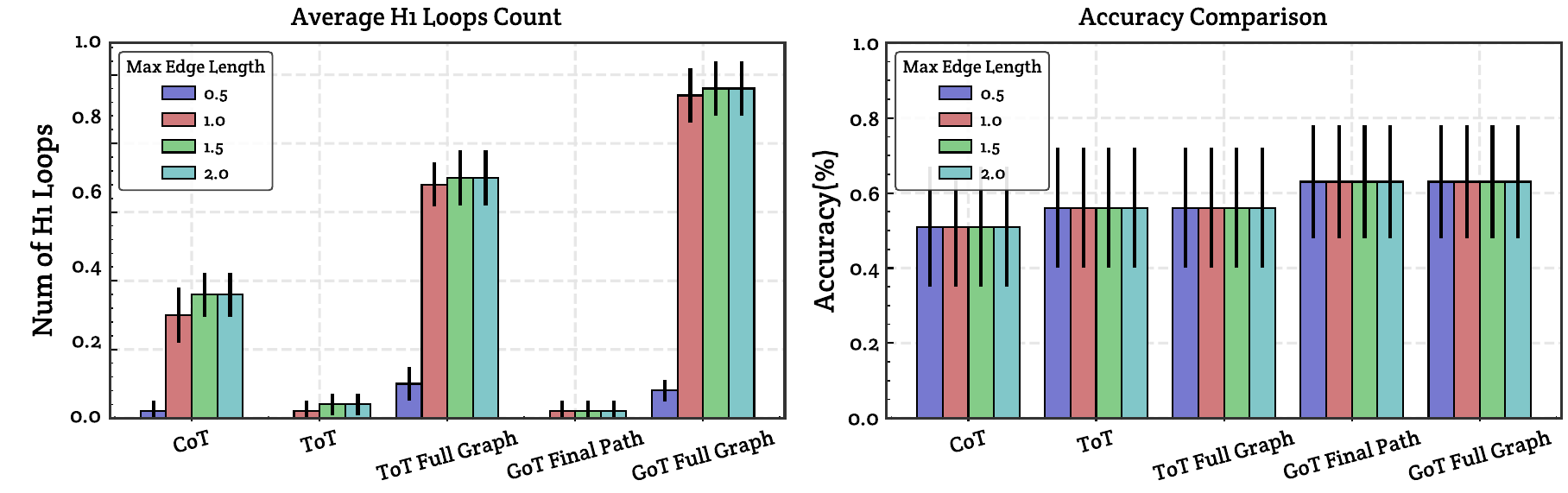}
	\caption{Relationship between $H_1$ components and the reasoning accuracy.}
    \label{fig:h1_reasoning_relation}
\end{figure*}

\noindent\textbf{$H_1$ Statistical Analysis.} 
\label{sec:h1_statistical_analysis}
For $H_1$ values, we differentiate between two perspectives: ``full path," which includes all nodes and edges in the complete graph generated during exploration, and ``final path," which only includes the path that leads to the correct answer. Notably, homology counts for $H_1$ in the full path are much higher, reflecting a complex topological structure with extensive connectivity, numerous cycles, and potential voids. This complexity results from the broad exploration phase, where various hypotheses and connections are tested, creating a rich semantic space that increases the chances of finding feasible solutions. 
In contrast, homology counts for the final path are lower, indicating a simpler structure with fewer components and cycles. In Fig.~\ref{fig:h1_reasoning_relation}, we see that longer-lasting $H_1$ values don’t necessarily represent meaningful semantic structures. These may just be closed arrangements of local vectors, not valid reasoning. This is consistent with insights from the heatmap, reinforcing the idea that effective reasoning requires a balance between exploration and focused synthesis.

\noindent\textbf{Reasoning Steps for Weekend Planning.}
\label{sec:cot_weekend_planning_vis}
In the concrete implementation of the visualization experiment in \S\ref{sec:2D_visualization_of_homological_structure}, the six reasoning steps are represented as nodes annotated with both a depth level and a branch index. The depth sequence \([0, 1, 1, 1, 2, 3]\) and branch sequence \([0, 0, 1, 2, 3, 3]\) jointly encode a tree structure with a single root, three alternative branches, and a subsequent merge leading to the final conclusion. Including: posing the core question of how to arrange a single weekend day efficiently without becoming exhausted; three candidate plans that emphasize, respectively, a ``study-focused,'' an ``exercise-focused,'' and a ``relaxation-focused'' schedule; and, finally, a mixed plan combining ``study + exercise + light leisure'' that is proposed after comparing the advantages and disadvantages of the three pure options. 
For each step, we obtain a semantic embedding and combine semantic and structural information into a single distance metric, assigning equal weight to the two components. The resulting representations are reduced to three dimensions using principal component analysis and then further projected to two dimensions for visualization. On top of these embedded points, we construct a Vietoris--Rips complex (using cosine distance for the semantic component and a radius threshold of \(\epsilon = 2\)) and compute homological statistics such as \(H_0\) and \(H_1\). In the visualization, the natural language description associated with each step is truncated to its first ten words to keep the labels readable in the two-dimensional layout.

\begin{lstlisting}[style=promptstyle]
steps = [
    # 0 root
    "Problem: I have one free weekend day and I want it to feel productive "
    "but not exhausting. I need to decide how to plan the day.",

    # 1 branch 0 -- study
    "Branch 0 -- Study-focused day: Spend most of the time studying "
    "(reading, taking notes). Pros: improves knowledge and long-term growth. "
    "Cons: can be tiring and may not feel restful.",

    # 2 branch 1 -- exercise
    "Branch 1 -- Exercise-focused day: Spend a large block of time exercising "
    "(running, gym). Pros: improves health and energy. "
    "Cons: limited direct progress on work or learning goals.",

    # 3 branch 2 -- relaxation
    "Branch 2 -- Relaxation-focused day: Mainly relax (watching shows, games, "
    "hanging out). Pros: very comfortable and reduces stress. "
    "Cons: may cause guilt later and little visible progress.",

    # 4 merge node -- compare branches
    "Compare the three options: A pure study day may lead to burnout and "
    "reduced efficiency. A pure exercise day helps body and mind but does not "
    "move key goals much. A pure relaxation day feels good but is likely to "
    "cause regret. A better idea is to combine them.",

    # 5 final conclusion
    "Conclusion -- Mixed plan: Spend 2--3 hours in the morning on focused study, "
    "1--2 hours in the afternoon on exercise, and keep the evening for light "
    "relaxation or social time. This balances progress, health, and recovery."
]
\end{lstlisting}

\subsection{Topological Reasoning Chains Extension Analysis}
\label{sec:tda_analyse}

This section presents a detailed description of mapping chains of thought into a semantic space within a topological data analysis framework. The framework not only preserves the temporal order and local coherence of the reasoning process, but also captures the structural organization of reasoning steps in semantic space—such as connectivity, clustering patterns, loop behavior, multi-path evidence, and structural breaks arising from semantic drift. By leveraging persistent homology and topological statistical indicators, we can quantify the structural robustness, redundancy, complexity, and stability of the logical backbone of reasoning chains across multiple scales.

\noindent\textbf{From Chains of Thought to Semantic Point Clouds: Formal Modeling.}
Suppose a chain of thought consists of $(n)$ reasoning steps, each being a sentence-level or paragraph-level text unit. We first define a semantic embedding function:
\begin{equation}
f: \mathcal{T} \rightarrow \mathbb{R}^d,
\end{equation}
where $\mathcal{T}$ denotes the set of all observable textual steps, and $f$ is a sentence embedding model. This maps the reasoning chain into:
\begin{equation}
X = \{x_1, x_2, \ldots, x_n\}, \qquad x_i = f(t_i) \in \mathbb{R}^d,
\end{equation}
endowing it with a semantic distance metric (typically cosine or Euclidean distance):
\begin{equation}
d(x_i, x_j) = \lVert x_i - x_j \rVert.
\end{equation}

In this way, the originally linear textual reasoning chain is transformed into: (1) a temporally ordered sequence of points, (2) a trajectory in a high-dimensional semantic space, (3) a point cloud from which topological structures can be constructed.

\begin{takeawaybox}
Extension: Simplices of different dimensions carry different reasoning implications.
\end{takeawaybox}
If semantic drift occurs within the reasoning chain (e.g., digressions, domain shifts, transitions from mathematical derivation to analogical explanation), it manifests as: (1) increased ``jump-like'' distances between consecutive points; (2) bends or deviations in the temporal trajectory; (3) the emergence of new connected components or $1$-cycles at the topological level.
These phenomena provide structural cues for subsequent topological analysis.

\noindent\textbf{VR Complex Construction: From Local Semantic Relations to Global Reasoning Organization.}
Given a threshold $\varepsilon$, the Vietoris--Rips complex is defined as:
\begin{equation}
VR_\varepsilon(X) 
= 
\{\sigma \subseteq X : d(x_i, x_j) \le \varepsilon,\; \forall x_i, x_j \in \sigma \}.
\end{equation}

\begin{takeawaybox}
Extension: Interpretations of Simplices of Different Dimensions.
\end{takeawaybox}

(1) 0-simplex (Point): Represents an independent reasoning step.
(2) 1-simplex (Edge): Represents a pair of reasoning steps that are semantically tightly related.
(3) 2-simplex (Triangle): Represents a locally dense cluster, such as a group of mutually related sub-steps.
(4) k-simplex (k-face): Represents a high-dimensional coherent reasoning cluster, such as the structure of a complex subproblem.

\begin{takeawaybox}
Extension: The VR complex reflects not only local semantic similarity but also the global organizational structure of reasoning when $\varepsilon$ becomes large:
\end{takeawaybox}

Geometric Intuition. (1) If the reasoning chain has a clear structure, isolated points will quickly merge into the main chain as $\varepsilon$ increases. (2) If the reasoning is disordered or noisy, the VR complex will exhibit numerous scattered small structures and complicated loops.

\noindent\textbf{Multi-Scale Filtration: A Multi-Resolution View of Reasoning Structure.}
We construct a filtration sequence:
\begin{equation}
VR_{\varepsilon_0}(X) 
\subseteq 
VR_{\varepsilon_1}(X) 
\subseteq 
\cdots 
\subseteq 
VR_{\varepsilon_m}(X).
\end{equation}

\begin{takeawaybox}
Extension: The scale ranges of the reasoning chain’s hierarchical structure correspond to different semantic interpretations of the reasoning process.
\end{takeawaybox}
(1) Small scales capture fine-grained local relations, often appearing as fragmented micro-steps.  
(2) Intermediate scales reveal the organization of semantic clusters, reflecting coherent reasoning modules.  
(3) Large scales expose the global topology of the reasoning process, such as overall connectedness or long-range semantic transitions.

\begin{takeawaybox}
Extension: There are three typical patterns of multi-scale structural evolution.
\end{takeawaybox}

(1) Rapid-Coalescence Chains (typical of high-quality reasoning):  
Small-scale structures appear fragmented, but they quickly merge into a single dominant backbone at intermediate scales.
(2) Persistently Fragmented Chains (indicative of disordered reasoning):  
Multiple branches remain even at large scales, suggesting long-term semantic drift or structural breaks.
(3) Multi-Cluster Structures (characteristic of multi-path reasoning):  
Several stable reasoning clusters emerge at intermediate scales, indicating that the model is simultaneously exploring multiple sub-paths.

\noindent\textbf{$H_0$: Connectivity, Structural Breaks, and Semantic Coherence Analysis.}
The 0-dimensional homology characterizes the number of connected components:
\begin{equation}
\beta_0(\varepsilon) = \text{connected components of } VR_\varepsilon(X).
\end{equation}

\begin{takeawaybox}
Extension: Different evolution patterns of $\beta_0(\varepsilon)$ correspond to different reasoning phenomena.
\end{takeawaybox}
(1) Local Noise vs.\ Global Coherence.  
If $\beta_0$ is large at small scales but decreases rapidly, the model exhibits fine-grained semantic variation while maintaining a coherent global theme.
If multiple components persist into medium or large scales, this indicates substantial structural breaks in the reasoning chain.
(2) Topological Signatures of Multi-Path Reasoning.  
Some reasoning strategies (e.g., parallel exploration of alternative solutions) produce multiple stable connected components at medium scales. 
If these components eventually merge at large scales, it suggests that the model has successfully integrated information from multiple reasoning paths.
(3) Topological Markers of Implicit Topic Drift.  
If a component remains isolated for a long range of scales without merging, it may indicate a “topic escape point,” signaling implicit semantic drift within the reasoning process.

\noindent\textbf{Homology Group Computation.}
\label{sec:homology_group_computation}
To analyze the topological features in the reasoning chain, we compute the homology group of each scale complex \(K_i\) in the filtration. Specifically, for each scale complex \(K_i\), we consider its \(k\)-dimensional homology group:
\begin{equation}
    H_k(K_i) = \frac{\ker \partial_k}{\operatorname{im} \partial_{k+1}},
    \label{eq:homology_group}
\end{equation}
where \(\partial_k\) is the boundary operator defined on the \(k\)-dimensional chain group \(C_k(K_i)\). The chain group \(C_k(K_i)\) consists of all possible \(k\)-dimensional simplices (such as points, line segments, triangles, etc.) in the complex \(K_i\), representing the geometric structure of the complex. The role of the boundary operator is to map a higher-dimensional simplex to its boundary, i.e., to extract the lower-dimensional part that constitutes the boundary of the high-dimensional geometric shape.

The set \(\ker \partial_k\) is the kernel, which includes all elements mapped to zero by the boundary operator. Intuitively, \(\ker \partial_k\) contains all closed \(k\)-chains, that is, \(k\)-dimensional chains whose boundary vanishes. These closed chains represent complete shapes in the topological space without any gaps or discontinuities.

The set \(\operatorname{im} \partial_{k+1}\) is the image, which consists of all \(k\)-chains generated as the boundary of \((k+1)\)-dimensional simplices. It represents the part of the complex that can be obtained from the higher-dimensional part via the boundary operator.

By computing the homology group
\(H_k(K_i) = \ker \partial_k / \operatorname{im} \partial_{k+1}\),
we extract the topological features of each complex. Specifically, \(H_0\) captures connected components, reflecting the semantic coherence of reasoning steps; \(H_1\) counts one-dimensional loops, indicating logical redundancy; while \(H_2\) and higher groups represent higher-order structures such as cavities, suggesting more complex reasoning patterns in the chain of thought.

\end{document}